\pdfoutput=1

\documentclass[11pt]{article}

\usepackage{authblk}

\usepackage{acl}

\usepackage{times}
\usepackage{booktabs}
\usepackage{latexsym}
\usepackage{paralist}
\usepackage[shortlabels]{enumitem}
\usepackage{graphicx}
\usepackage{todonotes}
\usepackage{colortbl}
\usepackage{multirow}
\usepackage{floatrow}
\usepackage{tabularx}
\usepackage{lscape}
\usepackage{xcolor}
\usepackage[nomessages]{fp}
\usepackage{calc}
\usepackage{tikz}
\usepackage{hyperref}
\usepackage{tabto}
\hypersetup{colorlinks=true, linkcolor=darkblue, filecolor=magenta, urlcolor=darkblue}
\usepackage{graphicx}

\usepackage[T1]{fontenc}

\usepackage[utf8]{inputenc}
\usepackage{fvextra, csquotes}

\usepackage{microtype}
\usepackage{subfiles}
\title{Multi-Source (Pre-)Training for Cross-Domain
Measurement, Unit and Context Extraction}

\author[1]{\textbf{Yueling Li}}
\author[2]{\textbf{Sebastian Martschat}}
\author[3]{\textbf{Simone Paolo Ponzetto}}
\affil[1]{ Knowledge Architecture, BASF Digital Solutions GmbH, 67061 Ludwigshafen, Germany}
\affil[2]{ Knowledge Architecture, BASF SE, 67056 Ludwigshafen, Germany}
\affil[3]{ Data and Web Science Group, University of Mannheim, 68131 Mannheim, Germany}
\affil[ ]{\texttt{\{yueling.li|sebastian.martschat\}@basf.com}}
\affil[ ]{\texttt{simone@informatik.uni-mannheim.de}}

\begin{document}
\maketitle
\begin{abstract}
We present a cross-domain approach for automated measurement and context extraction based on pre-trained language models. We construct a multi-source, multi-domain corpus and train an end-to-end extraction pipeline. We then apply multi-source task-adaptive pre-training and fine-tuning to benchmark the cross-domain generalization capability of our model. Further, we conceptualize and apply a task-specific error analysis and derive insights for future work. Our results suggest that multi-source training leads to the best overall results, while single-source training yields the best results for the respective individual domain. While our setup is successful at extracting quantity values and units, more research is needed to improve the extraction of contextual entities. We make the cross-domain corpus used in this work available online\footnote{\url{https://github.com/liy140/multidomain-measextract-corpus}}.
\end{abstract}

\section{Introduction}

Numeric components such as counts, measurements and are crucial information for researchers across various disciplines.
An automatic system for the extraction of numerical measurements (e.g. 10 \%) and related \textit{"contextual"} information such as measurands (e.g., concentration)  and measured entities (e.g., chemical solution) for scientific literature can aid in the efficient construction of knowledge bases \citep{Court.2018} for downstream research tasks.

Ideally, the system should be able to handle multiple subject domains or even unseen domains, as relying on multiple specialized systems is inefficient and sometimes infeasible: For instance, each specialized model requires dedicated training and deployment resources. Further, the target-domain cannot always be known at inference time. 

\paragraph{Related Work.}
Most existing work tackle domain-specific measurement extraction problems. In the bio-medical field alone, there
exist several publications that specialize in one particular sub-field and text genre, for instance food laboratory tests \citep{Kang.2013}, narrative radiology
reports \citep{Sevenster.2015} or diabetes test trials \citep{Hao.2016}. Many topic-specific systems are not designed as sole measurement extractors, additionally
offering comprehensive knowledge base construction or text processing functionalities \citep{Swain.2016, Dieb.2015, Epp.25.03.2021, Lentschat.06302020, friedrich-etal-2020-sofc}. Due to their specialization, it is difficult to apply them to novel domains without
considerable adaption effort.

A few domain-independent systems have been developed, but they either offer limited context extraction capabilities \citep{SoumiaLiliaBerrahou.2013, Mundler.2021} or lack concrete definitions of the extracted contextual entity types \citep{Foppiano.09232019, Hundman.2017b}. Moreover, for these systems, no study of cross-domain generalization capabilities has been performed.

\citet{harper-etal-2021-semeval}'s SemEval Task represents a key milestone for the progress of domain-independent measurement extraction research. 25 teams participated in the competition, thereby introducing the task, which has before mostly been solved through traditional NLP methods, to a wide range of recent architectures, among others also the now dominantly used pre-trained language model architecture \citep{Devlin.2019}.
Moreover, \citet{harper-etal-2021-semeval} define the task in a domain-agnostic manner and provide an annotated multi-domain measurement extraction corpus with data sourced from the OA-STM Corpus \citep{ElsevierLabs.2015}. However, due to its small data size (295 paragraphs), the corpus is not sufficient on its own for studying cross-generalization effects.

\paragraph{Contributions.} To address the research gaps mentioned above, i.e. the need for cross-domain generalization as an evaluation criterion for the measurement and context extraction task, we aim to build a \textit{cross-domain} measurement, unit \textit{and} context extraction system. We make the following contributions:
\begin{compactitem}
    \item To facilitate multi-domain training, we expand the corpus published by \citet{harper-etal-2021-semeval}, creating a multi-domain, multi-source corpus for measurement, unit, and context extraction including two additional source domains.
    \item We construct an end-to-end model pipeline based on pre-trained language models \cite{Devlin.2019} and achieve state-of-the-art performance comparable to the first placed MeasEval team \citep{davletov-etal-2021-liori-semeval}.
    \item We study the effect of (a) adaptive intermediate \textit{pre-training} \citep{Gururangan.2020} and (b) multi-source \textit{fine-tuning} \citep{Zhao.26.02.2020}  on cross-domain generalization. For (a) we apply full intermediate pre-training and adapter-based pre-training \citep{Hung.16.10.2021, Houlsby.2019} using a curated multi-domain task-adaptive pre-training corpus \citep{Gururangan.2020}. For (b), we experiment with different pooled combinations of \textit{fine-tuning} domains.
    \item Finally, we carry out a task-specialized error analysis using entity-level analysis methods adapted from \citet{Fu.2020} to determine concrete error sources for well-grounded model improvement.
\end{compactitem}
In the following sections, we explicate our corpus construction approach (§\ref{sec:corpus}), the model architecture (§\ref{sec:model}), and domain adaption methods (§\ref{sec:adaption}). Finally, we present our experimental results (§\ref{sec:results}) coupled with the error analysis (§\ref{sec:error_analysis}) and a concluding discussion (§\ref{sec:conc}).

\section{A New Multi-Domain Corpus for Measurement Extraction}
\label{sec:corpus}
In this section we describe the creation of a multi-domain corpus for measurement and context extraction. This will enable the investigation of cross-domain prediction performance.

\subsection{Data Model and Source Corpora} \label{subsec:corpus-data_model}
\begin{table*}[!ht]
\centering
\resizebox{0.8\textwidth}{!}{%
\begin{tabular}{@{}l|ccccc@{}}
\toprule
\textbf{Corpus} &
  \multicolumn{1}{l}{\textbf{Q}} &
  \multicolumn{1}{l}{\textbf{U}} &
  \multicolumn{1}{l}{\textbf{ME}} &
  \multicolumn{1}{l}{\textbf{MP}} &
  \multicolumn{1}{l}{\textbf{R}} \\ \midrule
\textbf{MeasEval Corpus }\citep{harper-etal-2021-semeval}                                                                               & x   & x    & x    & x   & x    \\
\begin{tabular}[c]{@{}l@{}}\textbf{Battery Materials Patents Corpus} (originally BASF internal dataset) \end{tabular}                     & (x) & x    & x    & x   & x    \\\begin{tabular}[c]{@{}l@{}}\textbf{Material Science Procedural Corpus} \citep{Mysore.2019}\end{tabular}                 & x   & x    & (x)  & (x) & (x)  \\

\begin{tabular}[c]{@{}l@{}}ChemDataExtractor Evaluation Corpus \citep{Swain.2016}\end{tabular}                & (x)   & x    & x    & x   & x \\
SOFC-Exp Corpus \citep{friedrich-etal-2020-sofc}                                                                              & x   & x    & (x)  & (x) & / \\\bottomrule
\end{tabular}
}\caption{\label{tab:data-candidates}Candidate datasets evaluated by data model components. Selected corpora are bolded. A full fit to the evaluation criterion is denoted with "x", a partial fit is indicated with "(x)" and a unrepresented concept is marked as "/". Q = Quantity Value, U = Unit, ME = MeasuredEntity, MP = MeasuredProperties, R = Relations.}
\end{table*}

The first step in corpus creation for measurement extraction is to decide on a data model that relates objects to be measured, values, and their context. We adapt the data model and terminology as proposed by \citet{harper-etal-2021-semeval}, excluding the "Qualifier" and "Modifier" classes to increase the candidate pool for corpus expansion.
\begin{figure}[!ht]
\centering
\includegraphics[width=0.8\textwidth]{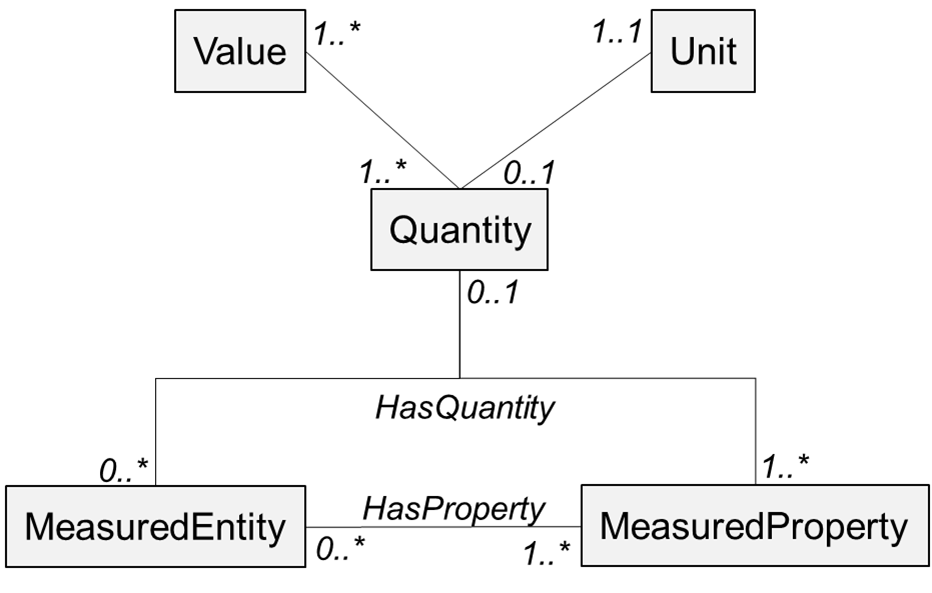}
\caption{\label{fig:corpus-data_model}Data model based on the MeasEval task definition \citep{harper-etal-2021-semeval}. Multiplicities between entities show the upper and lower bounds of entities for each relationship, i.e. 0..1 = zero or at most one, 0..* = zero or more, 1..1 = exactly one, 1..* = one or more.}
\end{figure}

\begin{figure}[!ht]
\centering
\includegraphics[width=1\textwidth]{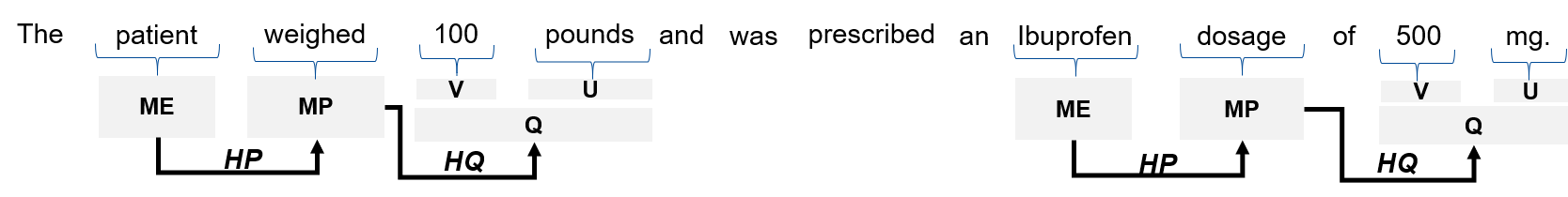}
\caption{Annotated example sentence. Additional abbreviations: HP - HasProperty, HQ - HasQuantity} 
\label{fig:corpus-example-sentence}
\end{figure}

Figure \ref{fig:corpus-data_model} presents the resulting adapted data model for our multi-source corpus: A \textbf{Quantity} (Q) is made up of one or more numeric Values (V) and optionally a \textbf{Unit} (U). A \textbf{MeasuredEntity} (ME) is the object, event or phenomenon, whose quantifiable property is measured. This would be the \textbf{MeasuredProperty} (MP), i.e., the measurand that can be attributed to the measured object. Figure \ref{fig:corpus-example-sentence} shows the annotation of an example sentence according to the presented data model. The extended data model definition can be found in Appendix \ref{appendix-data-model}.

\noindent

For selecting the source corpora, we compiled a candidate pool consisting of three datasets from related work and one additional patent dataset created by the chemical company BASF SE.
We then evaluated candidate datasets with respect to compatibility with the data model. Table \ref{tab:data-candidates} presents the evaluation summary. As such the resultant corpus is comprised of the MeasEval corpus \citep{harper-etal-2021-semeval}, the Battery Material Patents (BM) dataset as well as the Material Science Procedural (MSP) corpus \citep{Mysore.2019}.

\subsection{Data Processing and Annotation}
We apply several processing steps to normalize each source corpus with respect to the measurement extraction data model of Figure \ref{fig:corpus-data_model}. 

\paragraph{MeasEval Corpus.} For the MeasEval corpus, in order to accommodate the limited input length of most pre-trained models, we split the paragraphs into sentences using spaCy\footnote{\url{https://spacy.io}}. We deal with particularities of scientific language, e.g., bibliographic references and abbreviations by applying custom segmentation rules.

\paragraph{BM Corpus.} The BM dataset describes and classifies information regarding entities and properties of battery materials from patent claims. The original annotations specify the patent type of a claim (e.g., material claim vs. process claim) as well as phrase-level entity and relation information across 15 entity types and 13 relation types (e.g., stirrer elements, complexants, main metals). The entity types \textit{Value}, \textit{Unit} and \textit{Property} can be directly mapped to entities defined in our data model, i.e. Quantity value (Q), Unit (U) and MeasuredProperty (MP) respectively. By contrast, there are multiple source entity types that can be mapped to the MeasuredEntity (ME) class. These are parsed through graph traversal: we follow the relations that are connected to \textit{Value} entities, we thereby find their respective U, MP and ME. We save each claim separately and do not apply additional segmentation measures to preserve the unique structure of the patent style.

\paragraph{MSP Corpus.}
The MSP Corpus comprises 230 articles describing material synthesis procedures (MIT Open Source License, \citealt{Mysore.2019}). Although the annotation scheme is comparable to the Battery Materials dataset, including entity types such as \textit{Material}, \textit{Operation}, \textit{Amount-Unit}, \textit{Synthesis-Apparatus}, \textit{Number} etc., a more complex mapping would be required to cover the various semantic structures present in this dataset. For this reason, we opted to manually re-annotate the data instead of relying on entirely automatic re-labeling. To reduce the annotation effort, automatic labels using the original annotation data were created. As with the BM corpus, we follow all entities related to the entity \textit{Number} and map them to our defined labels (see \ref{appendix-data-msp-relabeling-mapping}).

The annotation process involved four non-native annotators from different scientific backgrounds and mixed genders.

We drafted a separate annotation guideline that (i) explained the task according to the MeasEval annotation guidelines and (ii) introduced dataset-specific instructions (Appendix \ref{appendix-reannotation-guidelines}). We re-annotate all samples of the validation articles (89 sentences) and test articles (129 sentences), and a subset of the training articles (860 sentences) to limit the annotation effort. We use the evaluation split for NER provided by \citet{Mysore.2019}.

An inter-annotator-agreement (IAA) study validated the reproducibility of our guidelines, producing substantial agreement scores. This is described further in Appendix \ref{appendix-iaa}. 

\paragraph{Final Corpus.}
The final multi-domain, multi-source corpus consists of the normalized version of the three corpora described above, totaling 2,389 sentences from scientific articles (MeasEval), 1,077 sentences from material synthesis procedures and 352 battery material patent claims (not segmented).
An overview of the final corpus is given in Appendix \ref{appendix-corpus-overview}.

\section{Extraction Architecture}
\label{sec:model}
\begin{figure*}[!htbp]
\centering
\includegraphics[width=0.7\textwidth]{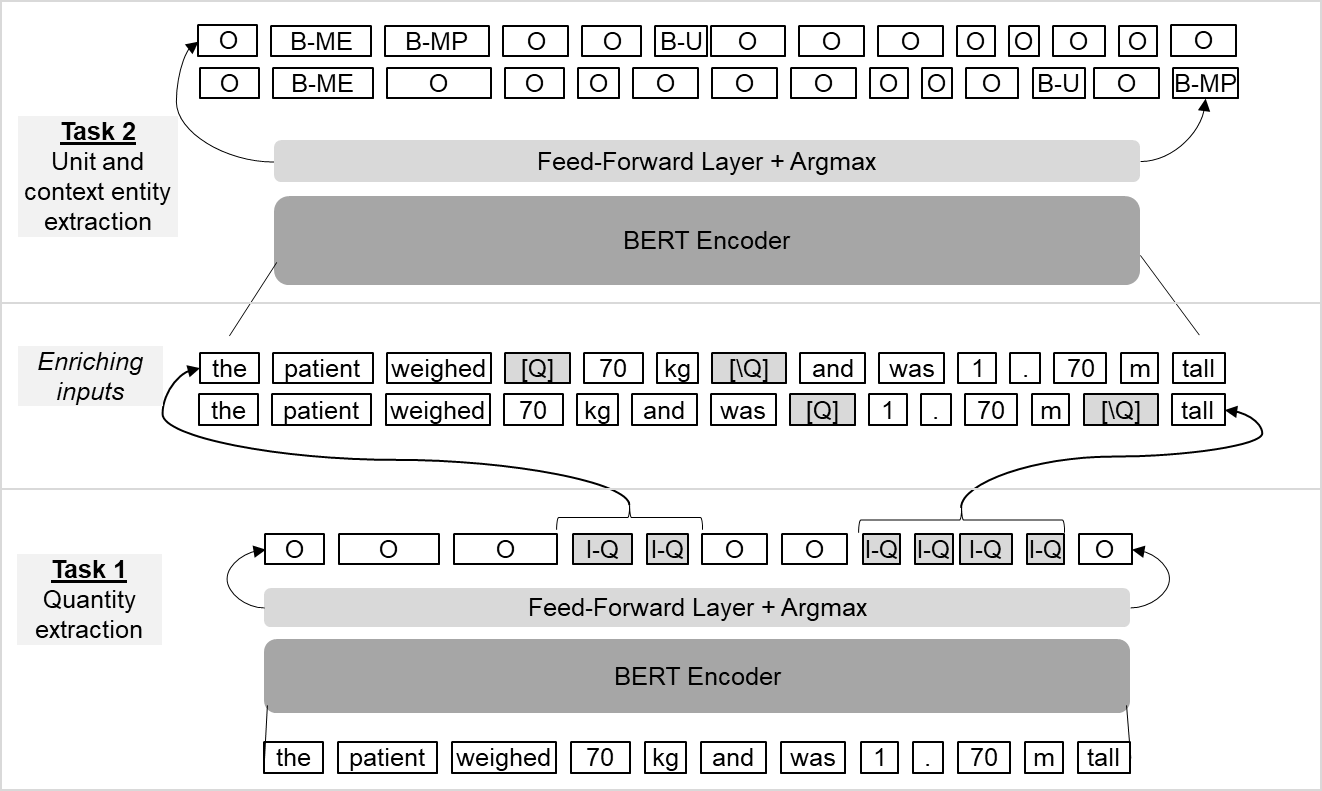}
\caption[Extraction flow]{Extraction flow, Q = Quantity, U = Unit, ME = MeasuredEntity, MP = MeasuredProperty} 
\label{fig:exp-extraction-flow}
\end{figure*}
We now describe our model setup which is designed to extract the entity and relation types described in the previous section.

We model the extraction as a two-step pipeline made up of two token-classification models, which we coin as \textbf{Task 1} and \textbf{Task 2}. We first extract all Quantities (Task 1) and then simultaneously predict U, ME and MP (Task 2) based on each extracted Quantity. This cascading setup resolves the relation extraction problem of assigning the context entities to the correct quantity span, as the data model allows for a deterministic, rule-based assignment of the relations between U, ME and MP (see \citet{Gangwar.03.04.2021, davletov-etal-2021-liori-semeval}).

Figure \ref{fig:exp-extraction-flow} shows the extraction flow based on an example sentence: The information from the first task is input into the second task through special tokens [Q] and [/Q] which we wrap around the identified Q spans (see also \citet{Gangwar.03.04.2021, davletov-etal-2021-liori-semeval}). For each identified Q, an enriched prediction sample is created, thereby allowing for overlapping entities and conditioning the unit and context entity extractor on one Q at a time. For Q extraction we use binary IO-tags \citep{liu-etal-2021-stanford}. For Task 2 we use the BIO-tagging scheme. To accommodate the tokens [Q] and [/\/Q] which signal the identified Q spans from Task 1, we add them to the models' vocabulary as special tokens, extending the embedding size by two. For training, we use cross-entropy loss over all classes and train Task 1 and Task 2 separately.

A drawback of this simple architecture is the fact that it cannot enforce the 1:1 relationships prescribed by the data model, since it is possible to predict more than one ME or MP. Further, we set the input sequence to the size of a single sentence to account for the one-sentence annotation window of the MSP and Battery Materials dataset.

\section{Domain Adaption and Generalization}
\label{sec:adaption}
\begin{figure}[!htbp]
\centering
\includegraphics[width=\textwidth]{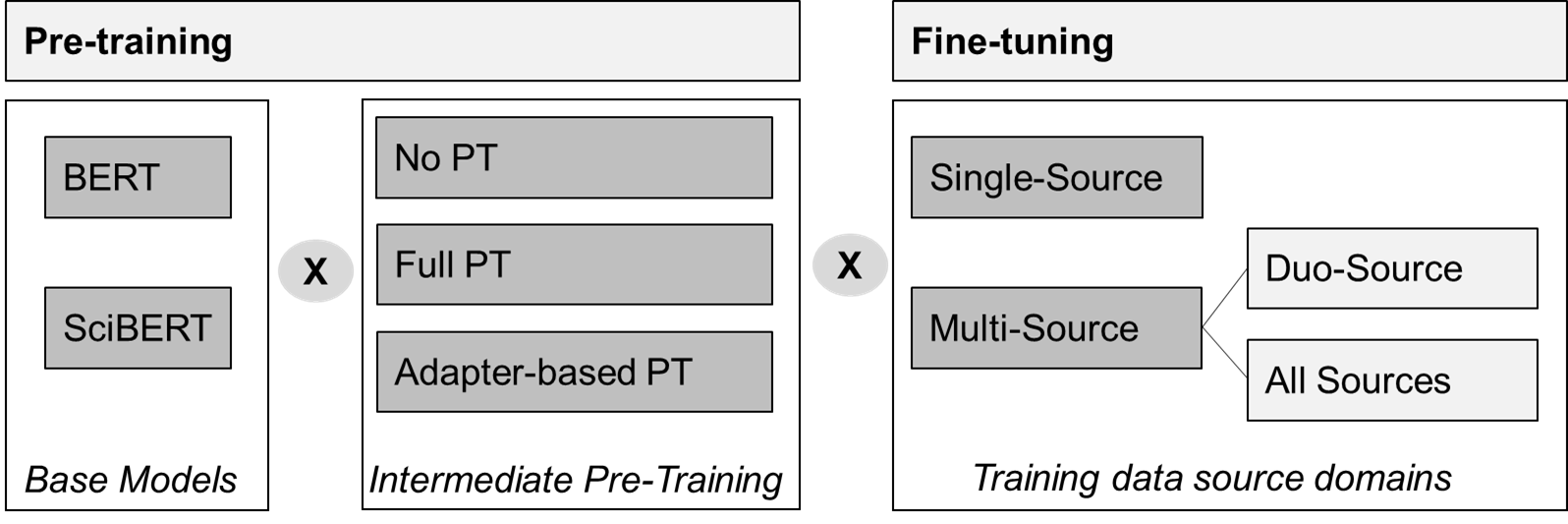}
\caption{Model configurations for domain adaptation and domain generalization by training phase.} 
\label{fig:method-domain-overview}
\end{figure}

We experiment with a) adaptive pre-training and b) multi-source fine-tuning. Figure \ref{fig:method-domain-overview} summarizes the applied methods and resulting model configurations. With the exception of the training setting with all sources, all shown configurations are applied to the models of both tasks.

\paragraph{Adaptive Pre-Training.} This setup comprises a combination of pre-trained base models and intermediate pre-training:
We use \textsc{Bert\textsubscript{Base}}\footnote{\url{https://huggingface.co/bert-base-uncased}} \citep{Devlin.2019} as the baseline model representing the canonical text domain, and SciBERT \footnote{\url{https://huggingface.co/allenai/scibert_scivocab_uncased}} \citep{Beltagy.2019}, which we expect to be more closely related to the domains of our measurement extraction corpus, because it was pre-trained from scratch on scientific articles.

We create intermediately pre-trained variants for each of the two BERT-models using task-adaptive pre-training (TAPT) \citep{Gururangan.2020}: we continue pre-training of the models on the unlabeled (training) data of our measurement extraction corpus. Thereby we aim to bring the models closer to the target domains of the task and induce increased task performance compared to the base models. We also apply adapter-based \citep{Houlsby.2019, Pfeiffer.2020} intermediate pre-training to compare full TAPT against to a more parameter efficient approach \citep{Kim.2021}.

As pre-training data, we create a "curated" \citep{Gururangan.2020} multi-source corpus comprised of the pooled data from all three training datasets. We enhance it with unlabled data from the same datasets to further increase the corpus size. As such we add all articles of the OA-STM dataset \citep{ElsevierLabs.2015}, the source on which the MeasEval annotations are based. We remove paragraphs which appear in the test and validation splits of the MeasEval data through fuzzy string matching \footnote{\url{https://github.com/seatgeek/thefuzz}}. Further, we add 1,128 Battery Materials claims which were excluded from the measurement extraction corpus due to lack of Quantity spans, and include the rest of the MSP data that was not re-annotated. This resulted in a pre-training corpus of approximately 630k words. 
\paragraph{Multi-Source Fine-Tuning.} 
To investigate the impact of multi-domain training, we also employ three experimental setups that are applied in the \textit{fine-tuning} stage of model training. In each setup a model is trained on one, two, or all domains and evaluated on all applicable target domains. The first set-up \textbf{single-source} uses only a single data set. To build our multi-source corpora we pool multiple data sources from related tasks \citep{Aue.2005, Zhao.26.02.2020}: As such, the second set-up \textbf{duo-source} uses the concatenation of two source datasets, e.g., BM + MeasEval, and the third set-up \textbf{all sources} uses all corpora. Due to considerable discrepancies between the Quantity annotation logic of the BM dataset and the other two datasets, no all sources setup was applied to Task 1.

\section{Experiments}
\label{sec:results}
We perform various experiments investigating the generalization capabilities of our system depending on data selection and domain adaption techniques. The implementation details can be found in Appendix \ref{appendix-implementation-details}.

\subsection{Evaluation and Scoring.} For comparing the predicted outputs to gold spans, we use the competition evaluation script provided by the MeasEval authors \citep{harper-etal-2021-semeval}, which is designed to jointly evaluate all sub-tasks by matching predicted Quantities to gold Quantities whilst taking into account the relationships between their respective contextual entities. To benchmark against the MeasEval competition results we report the competition metric \textit{Overlap F1} (see \citealt{harper-etal-2021-semeval}). For all other results, we report the traditional token-based strict F1, which is frequently used to evaluate NER and sequence-tagging tasks (cf.\ \citealt{Fried.09.07.2019, Swain.2016}). To this end, we adapt the MeasEval evaluation script by including nervaluate's \footnote{\url{https://github.com/MantisAI/nervaluate}} strict F1 implementation. %

\begin{table}[!t]
\resizebox{\textwidth}{!}{%
\begin{tabular}{@{}c|c|c|l|l||lll@{}}
\toprule
\multicolumn{1}{l}{Training} &
  \multicolumn{1}{l}{Source} &
  \multicolumn{1}{l}{} &
   &
  \multicolumn{1}{c}{Task 1} &
  \multicolumn{3}{l}{Task 2} \\ \cmidrule{5-8}
\multicolumn{1}{l}{Mode} &
  \multicolumn{1}{l}{Domain} &
  \multicolumn{1}{l}{Model} &
  PT Setup &
  \multicolumn{1}{c}{Q} &
  \multicolumn{1}{c}{U} &
  \multicolumn{1}{c}{ME} &
  \multicolumn{1}{c}{MP} \\\midrule
 &
   &
   &
  No PT &
  0.671 &
  0.96 &
  0.448 &
  0.473 \\
 &
   &
   &
  Full PT &
  0.702 &
  0.963 &
  0.424 &
  0.446 \\
 &
   &
  \multirow{-3}{*}{BERT} &
  Adpt. PT &
  0.688 &
  0.932 &
  0.388 &
  0.403 \\\cmidrule{3-8}
 &
   &
   &
  No PT &
  0.721 &
  0.972 &
  0.501 &
  0.522 \\
 &
   &
   &
  Full PT &
  0.719 &
  0.961 &
  0.491 &
  0.508 \\
 &
  \multirow{-6}{*}{\begin{tabular}[c]{@{}c@{}}Meas\\ Eval\end{tabular}} &
  \multirow{-3}{*}{SciBERT} &
  Adpt. PT &
  0.715 &
  0.952 &
  0.431 &
  0.425 \\\cmidrule{2-8}
 &
   &
   &
  No PT &
  0.632 &
  0.968 &
  0.452 &
  0.445 \\
 &
   &
   &
  Full PT &
  0.634 &
  0.956 &
  0.437 &
  0.461 \\
 &
   &
  \multirow{-3}{*}{BERT} &
  Adpt. PT &
  0.607 &
  0.946 &
  0.392 &
  0.421 \\\cmidrule{3-8}
 &
   &
   &
  No PT &
  0.667 &
  0.964 &
  0.456 &
  0.502 \\
 &
   &
   &
  Full PT &
  0.667 &
  0.965 &
  0.45 &
  0.5 \\
 &
  \multirow{-6}{*}{MSP} &
  \multirow{-3}{*}{SciBERT} &
  Adpt. PT &
  0.665 &
  0.952 &
  0.393 &
  0.491 \\ \cmidrule{2-8}
 &
   &
   &
  No PT &
  0.29 &
  0.865 &
  0.125 &
  0.216 \\
 &
   &
   &
  Full PT &
  0.285 &
  0.828 &
  0.148 &
  0.297 \\
 &
   &
  \multirow{-3}{*}{BERT} &
  Adpt. PT &
  0.315 &
  0.81 &
  0.119 &
  0.238 \\\cmidrule{3-8}
 &
   &
   &
  No PT &
  0.386 &
  0.75 &
  0.235 &
  0.355 \\
 &
   &
   &
  Full PT &
  0.357 &
  0.755 &
  0.21 &
  0.329 \\
\multirow{-18}{*}{\begin{tabular}[c]{@{}c@{}}Single-\\ source\end{tabular}} &
  \multirow{-6}{*}{BM} &
  \multirow{-3}{*}{SciBERT} &
  Adpt. PT &
  0.222 &
  0.559 &
  0.143 &
  0.263 \\\midrule
 &
   &
   &
  No PT &
  0.71 &
  0.968 &
  0.477 &
  0.538 \\
 &
   &
   &
  Full PT &
  0.726 &
  \textbf{0.974} &
  0.508 &
  0.496 \\
 &
   &
  \multirow{-3}{*}{BERT} &
  Adpt. PT &
  \textbf{0.732} &
  / &
  / &
  / \\ \cmidrule{3-8}
 &
   &
   &
  No PT &
  \textbf{0.739} &
  0.969 &
  \textbf{0.534} &
  \textbf{0.589} \\
 &
   &
   &
  Full PT &
  0.721 &
  \textbf{0.975} &
  \textbf{0.523} &
  0.557 \\
 &
  \multirow{-6}{*}{\begin{tabular}[c]{@{}c@{}}MSP+\\ Meas\\ Eval\end{tabular}} &
  \multirow{-3}{*}{SciBERT} &
  Adpt. PT &
  \textbf{0.727} &
  / &
  / &
  / \\ \cmidrule{2-8}
 &
   &
  BERT &
  No PT &
  / &
  0.968 &
  0.401 &
  0.441 \\
 &
   &
   &
  Full PT &
 /  &
  0.967 &
  0.439 &
  0.474 \\\cmidrule{3-8}
 &
   &
  SciBERT &
  No PT &
  / &
  0.97 &
  0.472 &
  0.54 \\
 &
  \multirow{-4}{*}{\begin{tabular}[c]{@{}c@{}}BM + \\ Meas\\ Eval\end{tabular}} &
   &
  Full PT &
 /  &
  0.965 &
  0.482 &
  0.537 \\ \cmidrule{2-8}
 &
   &
   &
  No PT &
  / &
  0.972 &
  0.455 &
  0.489 \\
 &
   &
  \multirow{-2}{*}{BERT} &
  Full PT &
  / &
  0.967 &
  0.417 &
  0.496 \\\cmidrule{3-8}
 &
   &
   &
  No PT &
  / &
  0.967 &
  0.435 &
  0.506 \\
\multirow{-14}{*}{\begin{tabular}[c]{@{}c@{}}Duo-\\ source\end{tabular}} &
  \multirow{-4}{*}{\begin{tabular}[c]{@{}c@{}}BM + \\ MSP\end{tabular}} &
  \multirow{-2}{*}{SciBERT} &
  Full PT &
  / &
  0.963 &
  0.448 &
  0.502 \\\midrule
 &
   &
   &
  No PT &
  / &
  \textbf{0.975} &
  0.479 &
  0.506 \\ 
 &
   &
   &
  Full PT &
  / &
  0.969 &
  0.46 &
  0.536 \\
 &
   &
  \multirow{-3}{*}{BERT} &
  Adpt. PT &
  / &
  0.969 &
  0.411 &
  0.484 \\ \cmidrule{3-8}
 &
   &
   &
  No PT &
  / &
  0.971 &
  0.512 &  
  \textbf{0.569} \\
 & 
   &
   &
  Full PT &
  / &
  0.972 &
  \textbf{0.524} &
  \textbf{0.593} \\ 
\multirow{-6}{*}{\begin{tabular}[c]{@{}c@{}}All \\ sources\end{tabular}} &
  \multirow{-6}{*}{\begin{tabular}[c]{@{}c@{}}MSP+\\ Meas\\ Eval+\\ BM\end{tabular}} &
  \multirow{-3}{*}{SciBERT} &
  Adpt. PT &
  / &
  0.956 &
  0.454 &
  0.535 \\ \bottomrule
\end{tabular}%
}\caption{Summary of the experiment results by task and extraction class. F1 scores are calculated based on the entire corpus. All setups were evaluated on the pooled test data from all domains.}
\label{tab:results-summary-multi-corpus}
\end{table}

\subsection{Results.} Table \ref{tab:results-summary-multi-corpus} shows a summary of the experiment scores evaluated on the entire multi-source corpus for Task 1 and Task 2. We observe that the models are rather accurate for the Q and U classes, while extraction performance for the contextual entities MP and ME is much lower. Below we study the results in more detail with regards to the influence of cross-domain fine-tuning and adaptive pre-training measures, and perform a dedicated analysis for the end-to-end performance of two pipeline compositions. The full result tables can be found in Appendix \ref{appendix-test-results}.
\paragraph{Adaptive Pre-training Experiments.}

\begin{figure}[t]
\centering
\includegraphics[width=0.6\textwidth]{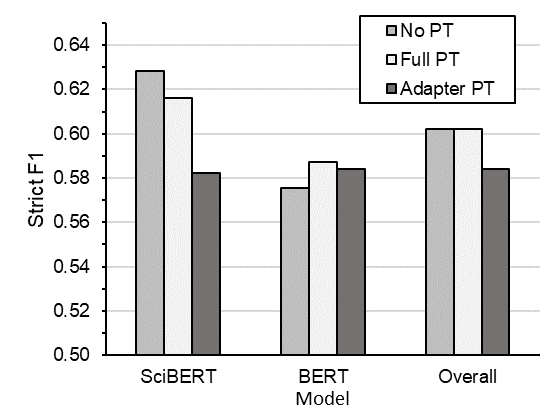}
\caption{Task 1 -- Avg.\ F1 score by model and pre-training setup.}
\label{fig:results-t1-model-pt}
\end{figure}

\begin{figure}[t]
\centering
\includegraphics[width=0.6\textwidth]{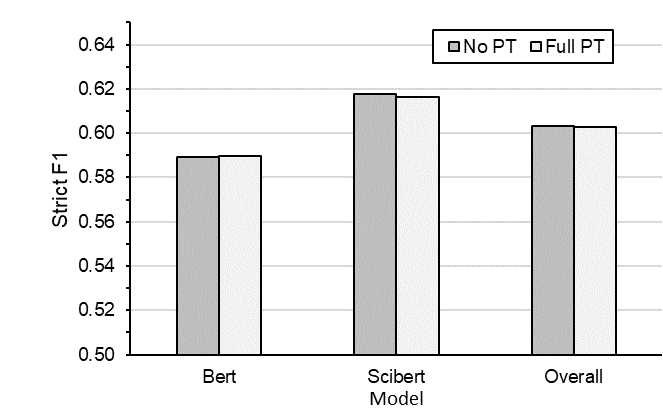}
\caption{Task 2 -- Avg.\ F1 score by model and pre-training setup.}
\label{fig:results-t2-model-pt}
\end{figure}

To study the effect of adaptive pre-training, we analyse the average F1 score by model type and task-adaptive pre-training setup (No PT vs.\ Full PT vs.\ Adapter PT):\\
\textbf{\textit{SciBERT vs. BERT}}: For both Task 1 and Task 2, we observe that both pre-trained and base SciBERT models achieve higher scores than their BERT counterparts (Figure \ref{fig:results-t1-model-pt}, Figure \ref{fig:results-t2-model-pt}).\\
\textbf{\textit{Full PT vs. No PT}}: When comparing different intermediate pre-training setups, we see no systematic patterns comparing Full PT and No PT performance. This is especially true for Task 2, where the average performance is quasi identical. Thus, we hypothesize that TAPT with limited pre-training data such as in our case is not sufficient for the more complex token classification task. \\
\textbf{\textit{Adapter PT}}: From Table \ref{tab:results-summary-multi-corpus} we see that adapter PT is often comparable or worse to Full PT and No PT. The performance drop is especially pronounced for the more challenging ME and MP class, while it remains comparable for the Q and U class.

\setlength\tabcolsep{4pt}
\begin{table*}[!ht]
\caption{End-to-end results using (strict) F1 measure.}
\label{tab:results-e2e}
\resizebox{\textwidth}{!}{%
\begin{tabular}{@{}ll|ccccc|ccccc|ccccc|ccccc@{}}\toprule
 &
   &
  \multicolumn{5}{c}{MeasEval} &
  \multicolumn{5}{c}{MSP} &
  \multicolumn{5}{c}{BM} &
  \multicolumn{5}{c}{Overall} \\\cmidrule{3-22}
\begin{tabular}[c]{@{}l@{}}Source   \\ domains\end{tabular} &
  \begin{tabular}[c]{@{}l@{}}Model Configuration \\ by Task (all SciBERT)\end{tabular} &
  All &
  Q &
  U &
  ME &
  MP &
  All &
  Q &
  U &
  ME &
  MP &
  All &
  Q &
  U &
  ME &
  MP &
  All &
  Q &
  U &
  ME &
  MP \\\midrule
\begin{tabular}[c]{@{}l@{}}MeasEval  \\ only\end{tabular} &
  \begin{tabular}[c]{@{}l@{}}T1: SciBERT Full PT; \\ T2: SciBERT Full PT\end{tabular} &
  \textbf{0.657} &
  \textbf{0.792} &
  \textbf{0.909} &
  \textbf{0.461} &
  0.442 &
  0.734 &
  0.915 &
  \textbf{0.971} &
  0.472 &
  0.550 &
  0.353 &
  0.379 &
  0.474 &
  0.251 &
  0.384 &
  0.602 &
  0.722 &
  0.844 &
  0.408 &
  0.449 \\\midrule
\begin{tabular}[c]{@{}l@{}}MSP \\ only\end{tabular} &
  \begin{tabular}[c]{@{}l@{}}T1: SciBERT No PT; \\ T2: SciBERT No PT\end{tabular} &
  0.569 &
  0.662 &
  0.870 &
  0.337 &
  0.370 &
  0.766 &
  0.915 &
  0.939 &
  0.538 &
  0.662 &
  0.411 &
  0.460 &
  0.468 &
  0.346 &
  0.388 &
  0.580 &
  0.675 &
  0.815 &
  0.386 &
  0.443 \\ \midrule
\begin{tabular}[c]{@{}l@{}}BM \\ only\end{tabular} &
  \begin{tabular}[c]{@{}l@{}}T1: SciBERT Full PT; \\ T2: SciBERT   Full PT\end{tabular} &
  0.310 &
  0.328 &
  0.602 &
  0.128 &
  0.198 &
  0.249 &
  0.304 &
  0.391 &
  0.136 &
  0.145 &
  0.443 &
  \textbf{0.505} &
  \textbf{0.524} &
  0.263 &
  \textbf{0.508} &
  0.328 &
  0.361 &
  0.538 &
  0.159 &
  0.283 \\ \bottomrule \bottomrule
Multi &
  \begin{tabular}[c]{@{}l@{}}T1: BM SciBERT Full PT \\       \& MSP+MeasEval Full PT; \\ T2: All Sources SciBERT No PT\end{tabular} &
  0.647 &
  0.782 &
  0.893 &
  0.445 &
  \textbf{0.456} &
  \textbf{0.776} &
  \textbf{0.930} &
  0.957 &
  \textbf{0.532} &
  \textbf{0.688} &
  \textbf{0.450} &
  \textbf{0.505} &
  0.519 &
  \textbf{0.354} &
  0.440 &
  \textbf{0.641} &
  \textbf{0.767} &
  \textbf{0.848} &
  \textbf{0.450} &
  \textbf{0.505}\\\bottomrule
\end{tabular}%
}
\end{table*}

\paragraph{Multi-Source Fine-Tuning Experiments.}
For the investigation of multi-source fine-tuning, we analyse the average F1 score by source domain and task (Table \ref{tab:results-t1-test-average}, Table \ref{tab:results-t2-test-average}): \\
\textbf{\textit{Cross-domain vs. in-domain}}: We observe that cross-domain fine-tuning is beneficial for overall generalization, while in-domain fine-tuning results in better in-domain performance.
An exception to that is the rather specialized, low-resource BM domain. Its best Task 2 performance is achieved in the MSP+MeasEval \textrightarrow BM cross-domain setup. Similarly, leaving BM out from multi-source training setups results in higher overall scores, suggesting that overly specialized training signals negatively impact generalization.
\begin{table}[!t]
\resizebox{\textwidth}{!}{%
\begin{tabular}{@{}llll|l@{}}
\toprule T1 classes: Q &  \multicolumn{4}{c}{Target domain} \\ \cmidrule{2-5}
 Source domain & MeasEval & MSP & BM & O \\ \midrule
MeasEval & 
\textbf{0.773} 
& \cellcolor[HTML]{F2F2F2}{\color[HTML]{808080} 0.847 }
& \cellcolor[HTML]{F2F2F2}{\color[HTML]{808080}  0.386}
& 0.703 \\
MSP & \cellcolor[HTML]{F2F2F2}{\color[HTML]{808080} 0.632} 
& 0.916 
& \cellcolor[HTML]{F2F2F2}{\color[HTML]{808080} 0.408 }
& 0.645 \\
BM 
& \cellcolor[HTML]{F2F2F2}{\color[HTML]{808080} 0.278 }
& \cellcolor[HTML]{F2F2F2}{\color[HTML]{808080} 0.215 }
& \textbf{0.467} 
& 0.309 \\
MSP+ MeasEval & 0.765 & \textbf{0.919} 
& \cellcolor[HTML]{F2F2F2}{\color[HTML]{808080} 0.424}
& \textbf{0.726} \\ \bottomrule
\end{tabular}%
}
  \caption{Task 1 -- Avg.\ F1 score by source domain. Grey cells indicate cross-domain prediction setups.}
\label{tab:results-t1-test-average}
\end{table}
\begin{table}[t]
\centering
\resizebox{\textwidth}{!}{%
\begin{tabular}{@{}llll|l@{}}
\toprule
  T2 classes: U, MP, ME              & \multicolumn{4}{c}{Target domain }                                                                                             \\ \cmidrule{2-5}
Source domain   & MeasEval                                             & MSP                                                  & BM    & Overall \\ \midrule
MeasEval &
  0.615 &
  \cellcolor[HTML]{F2F2F2}{\color[HTML]{808080} 0.655} &
  \cellcolor[HTML]{F2F2F2}{\color[HTML]{808080} 0.656} &
  0.631 \\
MSP &
  \cellcolor[HTML]{F2F2F2}{\color[HTML]{808080} 0.551} &
  \textbf{0.748} &
  \cellcolor[HTML]{F2F2F2}{\color[HTML]{808080} 0.661} &
  0.619 \\
BM &
  \cellcolor[HTML]{F2F2F2}{\color[HTML]{808080} 0.362} &
  \cellcolor[HTML]{F2F2F2}{\color[HTML]{808080} 0.365} &
  0.623 &
  0.400 \\
BM+MeasEval     & 0.612                                                & \cellcolor[HTML]{F2F2F2}{\color[HTML]{808080} 0.642} & 0.656 & 0.626   \\
BM+MSP          & \cellcolor[HTML]{F2F2F2}{\color[HTML]{808080} 0.556} & 0.743                                                & 0.665 & 0.621   \\
MSP+MeasEval &
  0.621 &
  0.744 &
  \cellcolor[HTML]{F2F2F2}{\color[HTML]{808080} \textbf{0.700}} &
  \textbf{0.664} \\
MSP+MeasEval+BM & \textbf{0.627}                                       & 0.741                                                & 0.657 & 0.661   \\ \bottomrule
\end{tabular}%
}
\caption[Task 2: Average F1 score aggregated by source domain]{Task 2 -- Average F1 score aggregated by source domain. Grey cells
indicate cross-domain prediction setups.}
\label{tab:results-t2-test-average}
\end{table} %

\paragraph{End-To-End Evaluation.}
Table \ref{tab:results-e2e} shows the resulting E2E performance for selected model configurations, which allows us to assess the error propagation of the the cascading task flow.

We apply two separate Task 1 models for the multi-source setup, as we have not trained a model using all three datasets for Task 1 due to diverging Quantity annotation styles between BM and the other two data sources. We observe that the best overall end-to-end performance is achieved in the multi-source scenario, caused by superior performance on the MSP and BM domains. For MeasEval, we notice that the Unit extraction scores remain relatively high given the 0.2 drop in Quantity extraction. Remarkably, we see that Unit extraction works better in the MeasEval\textrightarrow MSP setup than in the in-domain setup.

\subsection{Comparison with MeasEval Leaderboard}
In Table \ref{tab:benchmark_measeval} we compare a single-source and a multi-source end-to-end setup against the highest-ranking team of the MeasEval competition. All models were selected based on the best strict F1 MeasEval target domain performance (as opposed to the overall performance) of the development data.

\begin{table}[t]
\centering
\resizebox{\textwidth}{!}{%
\begin{tabular}{@{}lccccccc@{}}
\toprule
Model & Q & U & ME & MP & HQ & HP & O \\ \midrule
\begin{tabular}[c]{@{}l@{}}1st place MeasEval \\ \citet{davletov-etal-2021-liori-semeval} \end{tabular} & 0.861 & 0.722 & \textbf{0.437} & \textbf{0.467} & \textbf{0.482}  & \textbf{0.318}  & \textbf{0.551} \\ \addlinespace
\begin{tabular}[c]{@{}l@{}}Single-source setup \\ (T1: MeasEval+SciBERT+Full PT;\\ T2: MeasEval+SciBERT+No PT) \end{tabular}             & \textbf{0.877} & \textbf{0.885} & 0.432 & 0.437 & 0.465 & 0.307 & 0.550 \\\addlinespace
\begin{tabular}[c]{@{}l@{}}Multi-source setup \\ (T1: MSP+MeasEval+SciBERT+Full PT;\\ T2: All sources+SciBERT+No PT)\end{tabular}       & 0.876 & 0.864 & 0.404 & 0.440 & 0.46 & 0.27 & 0.533 \\ \bottomrule
\end{tabular}%
}
\caption[MeasEval benchmark]{Benchmarking against MeasEval leaderboard's top team, scores correspond to MeasEval's competition scoring overlap F1.}
\label{tab:benchmark_measeval}
\end{table}

Our single-source setup performs on par with the winning team from \citet{davletov-etal-2021-liori-semeval}, showing superior scores for the Q and U classes, comparable scores for the ME class, and inferior scores for the MP class and the relation classes. As such we achieve competitive results with an arguably simpler model setup: \citet{davletov-etal-2021-liori-semeval}'s quantity extraction model is based on an ensemble of multiple LUKE models and entity-aware self-attention \citep{Yamada.2020}. Further, they use XLM-RoBERTa-large for unit and context span extraction and apply multi-task learning with parallel task-specific layers for each entity type. 
Moreover, we work with a smaller input context of one single sentence, while the winning team applies a data augmentation technique, increasing the available context
\citep{davletov-etal-2021-liori-semeval}.
However, we point out that their model learns far more entity types at the same time (seven in total), as we only work with a subset of the MeasEval task definition.

\section{Error Analysis}
\label{sec:error_analysis}

To better understand the challenges of the task and deficiencies of our system, we perform an in-depth error analysis. We analyze error sources on a fine-grained entity level. To prevent the leakage of test data knowledge, we apply all error analysis methods on the development portion of the corpus using our best development model setup (Task 1: BM+SciBERT+Full PT \& Duo-Source(MSP+MeasEval)+Full PT; Task 2: All Sources+SciBERT+No PT). Due to the relatively high scores for the Unit class, we focus the analysis on the Q, ME and MP classes.

\paragraph{Entity Data Attributes.}
To detect model weaknesses related to the properties of entity spans, we draw on the notion of \textit{data attributes} as defined by \citet{Fu.2020}: These are \textit{"[...] values which characterize the properties of an entity that may be correlated with the NER performance."} (p. 6059). These values can be related to characteristics of the entity's surface string (e.g., entity length) or its surrounding context (e.g., sentence length).
We analyse the following attributes:\footnote{We use Huggingface’s BertTokenizerFast based on SciBERT vocabulary for tokenization based attributes \url{https://huggingface.co/docs/transformers/model_doc/bert}}:
\newlength\WIDTHOFBAR
\setlength\WIDTHOFBAR{1cm}

\newlength{\mylength}

\newcommand{\blackwhitebar}[2]{%
 \FPeval{\result}{round(#1/#2, 1)}
  {\color{black!100}\rule{\result cm}{8pt}}{\color{black!30}\rule{\WIDTHOFBAR - \result cm}{8pt}} #1 }
 
\begin{table*}[!htbp]
\resizebox{\textwidth}{!}{%
\begin{tabular}{@{}cl||lll|lll|lll||lll@{}}
\toprule
\multicolumn{1}{l}{} &
   &
  \multicolumn{3}{c}{$eLen$} &
  \multicolumn{3}{c}{{$eDen$}} &
  \multicolumn{3}{c}{{$qDist$}} &
  \multicolumn{3}{c}{{Match Type Count}} \\ \midrule
\multicolumn{1}{l}{{Class}} &
  Match Type &
  MeasEval &
  MSP &
  BM &
  MeasEval &
  MSP &
  BM &
  MeasEval &
  MSP &
  BM &
  MeasEval &
  MSP &
  BM \\ \midrule
 &
  match &
 \blackwhitebar{3.9}{5.5} &
  \blackwhitebar{3.0}{15.1} &
  \blackwhitebar{8.9}{16.9} &
  \blackwhitebar{0.130}{0.184} &
   \blackwhitebar{0.202}{0.220}  &
  \blackwhitebar{0.073}{0.108}  &
  \cellcolor[HTML]{C0C0C0} &
  \cellcolor[HTML]{C0C0C0} &
  \cellcolor[HTML]{C0C0C0} &
  223 &
  179 &
  86 \\
 &
  partial &
 \blackwhitebar{5.5}{5.5} &
  \blackwhitebar{8.9}{15.1} &
  \blackwhitebar{8.6}{16.9} &
  \blackwhitebar{0.153}{0.184} &
  \blackwhitebar{0.220}{0.220}  &
  \blackwhitebar{ 0.065}{0.108}  &
  \cellcolor[HTML]{C0C0C0} &
  \cellcolor[HTML]{C0C0C0} &
  \cellcolor[HTML]{C0C0C0} &
  57 &
  14 &
  34 \\
 &
  spurious &
 \blackwhitebar{1.6}{5.5} &
  \blackwhitebar{2.8}{15.1} &
  \blackwhitebar{3.7}{16.9} &
  \blackwhitebar{0.184}{0.184} &
  \blackwhitebar{0.215}{0.220}  &
  \blackwhitebar{0.108}{0.108} &
  \cellcolor[HTML]{C0C0C0} &
  \cellcolor[HTML]{C0C0C0} &
  \cellcolor[HTML]{C0C0C0} &
  5 &
  6 &
  3 \\
\multirow{-4}{*}{{Quantity}} &
  missing &
 \blackwhitebar{1.8}{5.5} &
  \blackwhitebar{2.5}{15.1} &
  \blackwhitebar{2.2}{16.9} &
  \blackwhitebar{0.030}{0.184} &
  \blackwhitebar{0.147}{0.220}  &
  \blackwhitebar{0.073}{0.108}  &
  \cellcolor[HTML]{C0C0C0} &
  \cellcolor[HTML]{C0C0C0} &
  \cellcolor[HTML]{C0C0C0} &
  24 &
  6 &
  19 \\ \midrule
 &
  match &
 \blackwhitebar{2.8}{5.5} &
  \blackwhitebar{3.7}{15.1} &
  \blackwhitebar{2.7}{16.9} &
  \blackwhitebar{0.126}{0.184} &
  \blackwhitebar{0.204}{0.220}&
   \blackwhitebar{0.070}{0.108} &
  \blackwhitebar{18}{167} &
  \blackwhitebar{22}{57} &
  \blackwhitebar{73}{127} &
  85 &
  119 &
  44 \\
 &
  partial &
 \blackwhitebar{5.4}{5.5} &
  \blackwhitebar{15.1}{15.1} &
  \blackwhitebar{2.8}{16.9} &
  \blackwhitebar{0.125}{0.184} &
  \blackwhitebar{0.198}{0.220}  &
 \blackwhitebar{0.093}{0.108}  &
  \blackwhitebar{18}{167} &
  \blackwhitebar{55}{57} &
  \blackwhitebar{39}{127} &
  45 &
  59 &
  40 \\
 &
  missing &
 \blackwhitebar{2.9}{5.5} &
  \blackwhitebar{14.0}{15.1} &
  \blackwhitebar{2.6}{16.9} &
 \blackwhitebar{0.169}{0.184}  &
  \blackwhitebar{0.184}{0.220}  &
  \blackwhitebar{0.056}{0.108}  &
  \blackwhitebar{45}{167} &
  \blackwhitebar{57}{57} &
  \blackwhitebar{127}{127} &
  149 &
  29 &
  38 \\
\multirow{-4}{*}{{\begin{tabular}[c]{@{}c@{}}Measured\\      Entity\end{tabular}}} &
  spurious &
 \blackwhitebar{2.2}{5.5} &
  \blackwhitebar{1.6}{15.1} &
  \blackwhitebar{1.9}{16.9} &
  \blackwhitebar{0.132}{0.184} &
  \blackwhitebar{0.173}{0.220}  &
  \blackwhitebar{0.057}{0.108} &
  \blackwhitebar{167}{167} &
  \blackwhitebar{46}{57} &
  \blackwhitebar{93}{127} &
  119 &
  38 &
  44 \\  \midrule
 &
  match &
 \blackwhitebar{1.9}{5.5} &
  \blackwhitebar{1.5}{15.1} &
  \blackwhitebar{3.3}{16.9} &
  \blackwhitebar{0.142}{0.184}   &
  \blackwhitebar{0.215}{0.220}  &
  \blackwhitebar{0.077}{0.108} &
  \blackwhitebar{10}{167} &
  \blackwhitebar{22}{57} &
  \blackwhitebar{17}{127} &
  71 &
  102 &
  90 \\
 &
  partial &
 \blackwhitebar{3.5}{5.5} &
  \blackwhitebar{3.6}{15.1} &
  \blackwhitebar{5.0}{16.9} &
  \blackwhitebar{0.139}{0.184}  &
  \blackwhitebar{0.195}{0.220} &
  \blackwhitebar{0.051}{0.108} &
  \blackwhitebar{22}{167} &
  \blackwhitebar{28}{57} &
  \blackwhitebar{30}{127} &
  38 &
  17 &
  25 \\
 &
  missing &
  \blackwhitebar{1.9}{5.5} &
  \blackwhitebar{2.2}{15.1} &
  \blackwhitebar{16.9}{16.9} &
  \blackwhitebar{0.177}{0.184}  &
  \blackwhitebar{0.220}{0.220} &
  \blackwhitebar{0.069}{0.108} &
  \blackwhitebar{40}{167} &
  \blackwhitebar{23}{57} &
  \blackwhitebar{37}{127} &
  70 &
  29 &
  8 \\
\multirow{-4}{*}{{\begin{tabular}[c]{@{}c@{}}Measured\\      Property\end{tabular}}} &
  spurious &
 \blackwhitebar{1.7}{5.5} &
  \blackwhitebar{1.4}{15.1} &
  \blackwhitebar{4.2}{16.9} &
  \blackwhitebar{0.124}{0.184}  &
  \blackwhitebar{0.187}{0.220}  &
  \blackwhitebar{0.076 }{0.108}&
  \blackwhitebar{70}{167} &
  \blackwhitebar{28}{57} &
  \blackwhitebar{10}{127} &
  112 &
  37 &
  21 \\ \bottomrule 
\end{tabular}}
\caption{Count and average entity attributes $sLen$, $eDen$, $qDist$ by domain, class and match type. Bar dimensions are scaled to each domain. }
\label{tab:entity-attribute-error-analysis}
\end{table*}

\begin{compactitem}
    \item  Entity length ($eLen$): The number of tokens in an entity.
    \item Sentence entity density ($eDen$): The number of entities in a sentence divided by the sentence length. Thus, paragraphs with multiple measurements and associated contextual entities will have a higher entity density than paragraphs with a single measurement.
    \item Gold quantity distance ($qDist$): The character-level span distance of the gold entity to its associated gold Q. $qDist$ only applies to the classes ME and MP. We have filtered out all cross-sentence entities for this attribute.
\end{compactitem}

\paragraph{Analysis.}
In our approach, we first calculate the described data attributes. For (partial) matches and missing predictions we base the calculation on the gold entity span, for spurious predictions we base the it on the predicted span. Then, we average the attributes by match type, i.e., match, partial (match), missing and spurious, to allow the comparison of attribute averages between matches and errors. The last three columns of Table \ref{tab:entity-attribute-error-analysis} show the distribution of match types by domain and entity class. We see that the number of matches is especially high for the Q class, while the number of errors is especially high in the ME class of the MeasEval and BM domains. The remaining columns show the grouped attribute averages by domain, entity class and match type. The bar charts indicate the relative magnitude of an attribute mean within one domain. We make the following observations:
\begin{compactitem}
    \item \textbf{eLen}: Partial matches occur particularly for longer entities. For instance, we observe that tokens such as brackets or dashes within a longer entity are often broken up and predicted with different labels (e.g., ”deionized (DI) water” is broken into ”deionized”, ”DI) water”). Further, spurious predictions are always relatively short, often shorter than the average $eLen$ of matches. For MSP the missing MEs have a high $eLen$, suggesting that the model has difficulties extracting longer phrases. The same phenomenon holds for the MPs of the BM domain.
    \item \textbf{eDen}: For both MeasEval and MSP spurious Qs are predicted for sentences with a lower entity density. Further, we observe that both missing MPs and MEs of the MeasEval data and missing Qs of the BM data appear in sentences with higher entity density. This implies that the model may 'overlook' entities when many potential entities are in one area.
    \item \textbf{qDist}: The model mainly struggles with long range dependencies for the ME class: $qDist$ shows the distance of the supposed gold ME to its root Q by match type. We see that our setup is good at predicting MEs that are close to their root Q, as $qDist$ is rather small for matches. However, for higher $qDist$ MEs the match type is often missing or spurious. This means that the model does not predict an ME at all (missing), or predicts a spurious one, probably closer to the root Quantity. This issue does not apply to MPs, as their $qDist$ is much lower on average.
\end{compactitem}

\section{Conclusion}
\label{sec:conc}
We have applied pre-trained language models to end-to-end measurement, unit and context extraction. 
While our setup exhibits good extraction performance for Quantities and Units, more research has to be done to improve the extraction of contextual entities. We have identified long-range dependencies of MEs as a particular error source. \\
\indent In terms of cross-domain generalization and multi-source training, multi-source training produced the best overall results, while single-source training often yielded the best results for the respective target domain. An exception to this was the small BM dataset, for which we observed the best unit and context extraction performance in the cross-domain prediction setting. This is an indicator for domain generalization, especially for low-resource domains. However, this needs to be confirmed in additional experiments with a dataset comprising even more domains. \\
\indent When comparing {adaptive pre-training methods}, the most consistent performance driver was the use of the SciBERT base model instead of the BERT base model. Further, we found adapter-based intermediate pre-training to be worse in most cases for both model types and tasks, which may be due to the task complexity. This theory is affirmed by the fact that we saw better adapter-based pre-training results for the simpler Quantity extraction.

Finally, we found non-conclusive results for the comparison of no pre-training versus full pre-training. The instability of results may be due to the limited size of our pre-training data, or the effect of catastrophic forgetting. Future work with a larger pre-training corpus may give clearer insights into this case. 

\section*{Limitations and Ethical Considerations}
We would like to discuss the following limitations and ethical considerations:

In this paper, we investigated the cross-domain extraction performance based on a multi-source corpus. Our working assumption is that this corpus represents enough variety to support such a claim. However, we point out that the corpus is biased towards English scientific and patent language, as well as the chemical / material science subject domain. 
Further, we remark that the subjects distribution itself is biased towards the BM and MSP datasets as the the more varied MeasEval dataset only contains few examples for each of its 10 subjects. Consequently, a balanced corpus should have a more even distribution of both subject domains and language domains by increasing the size of the currently underrepresented domains and ideally including data from more than only the English language.

Further, despite having substantial IAA scores for the re-annotation of the MSP corpus, we often perceived the task as difficult and ambiguous and felt the limitations of only having two contextual entities, instead of the three as proposed by \citet{harper-etal-2021-semeval}. Yet, the low IAA score (0.334) for the excluded Qualifier entity suggests that including it may not have eased the task. Hence, it may be valuable to further the study of how the measurement extraction problem can be modelled to resolve some of the ambiguities for context extraction.

Finally, while we tried to stay as closely to the original annotation guidelines as proposed by \citet{harper-etal-2021-semeval} as possible (with the exception of the two cases explicated in Appendix \ref{appendix-reannotation-guidelines}, there is a high likelihood of annotation drift. The re-annotators of the MSP corpus were not involved in the original MeasEval annotation procedure and it is possible that the interpretation of the annotation guidelines was slightly different at places than the authors have originally intended. Our adaption of the annotation guidelines can be found at the end of this paper (Appendix \hyperref[appendix-guidelines]{J}).

\bibliography{custom}
\bibliographystyle{acl_natbib}
\clearpage

\appendix

\label{sec:appendix}
\section*{Appendix}
\section{Extended Description of the Data Model} \label{appendix-data-model}
Below we explicate each of the entity types, relations and their associated cardinalities of our data model which are largely based on \citet{harper-etal-2021-semeval}'s definitions.

\begin{enumerate}[1.]
    \item \textbf{Entities} define \textit{what} is to be extracted: \begin{compactitem}
        \item \textbf{Quantity (Q)}: A quantity is made up of a) one or more numeric \textbf{values} or counts signifying amounts or measurements and optionally b) a \textbf{Unit} (U) indicating the magnitude of the values. According to the MeasEval annotation guidelines, values and units are annotated in one span where possible. Contiguous values of a range or a list belong to the same Quantity span (e.g. "Possible beverage sizes are \textit{200, 300 or 400 ml}").
        \item \textbf{MeasuredEntity (ME}: A measured entity is the \textit{object}, event or phenomenon, whose quantifiable property is measured.
        \item \textbf{MeasuredProperty (MP)}: A measured property is a quantifiable property of the MeasuredEntity, i.e., the \textit{measurand} that can be attributed to the measured \textit{object}.
    \end{compactitem}
    \item \textbf{Relations} define \textit{how many} entities can be extracted and how they \textit{relate} to each other: \begin{compactitem}
        \item \textbf{HasQuantity (HQ)}: This relationship links the context entities to their respective quantities. This relation can be drawn from a MeasuredEntity to the Quantity, if no associated MeasuredProperty exists. Otherwise, it is drawn from the MeasuredProperty to the Quantity. The cardinalities of this relationship show that there can be at most one HasQuantity relation for any Quantity span, whereas any MeasuredProperty or MeasuredEntity can be linked to multiple Quantity spans. Consequently, there can be at most one MeasuredProperty and one MeasuredEntity linked to any Quantity span. Consider the sentence \textit{"The book was 600 pages long and weighed 0.5 kg."}. Here, the MeasuredEntity \textit{"book"} can be linked to two Quantity spans.
        \item \textbf{HasProperty (HP)}: This relation shows which MeasuredProperties can be attributed to a MeasuredEntity. While there can be MeasuredEntities without associated MeasuredProperties, the MeasEval data scheme prescribes that there must be a MeasuredEntity for any MeasuredProperty.
    \end{compactitem}
\end{enumerate}

\section{MSP automatic label suggestions for re-annotation}\label{appendix-data-msp-relabeling-mapping}
\begin{table}[!ht]
\centering
\resizebox{\textwidth}{!}{
\begin{tabular}{@{}lll@{}}
\toprule
\textbf{Unit}  & \textbf{MeasuredEntity}    & \textbf{MeasuredProperty} \\ \midrule
Amount-Unit    & Material                   & Apparatus-Property-Type   \\
Property-Unit  & Material-Descriptor        & Amount-Misc               \\
Apparatus-Unit & Nonrecipe-Material         & Property-Type             \\
Condition-Unit & Synthesis-Apparatus        & Condition-Misc            \\
               & Apparatus-Descriptor       & Condition-Type            \\
               & Characterization-Apparatus &                           \\
               & Property-Misc              &                           \\ \bottomrule
\end{tabular}}
\caption{Mapping of MSP entities to measurement extraction entities}
\end{table}

\section{Annotation Guidelines for the Re-Annotation of the MSP dataset}\label{appendix-reannotation-guidelines}
We provide the complete annotation guidelines for the re-annotation of the MSP dataset as at the end of this document (see Appendix \hyperref[appendix-guidelines](J).

Below we explicate specific re-annotation guidelines, diverging from the original measurement extraction guidelines provided by \citet{harper-etal-2021-semeval}.

\paragraph{Specific re-annotation guidelines} Over the course of the annotation procedure, the annotators have agreed on additional guiding principles to better capture the relationships between measurements and entities in the context of experiments as described by the synthesis procedures:

\begin{description}
    \item[Material-centered annotation] As a general rule, we prioritize the annotation of experimental participants over other attributes of the experimental procedure. In the sentence "The mixture of elements were heated in evacuated quartz ampoules at 1220 K [...]", we would annotate the "mixture of elements" as the MeasuredEntity of the Quantity "1220 K" as opposed to the "evacuated quartz ampoules".
    \item[Experimental conditions] Temperatures, times or rates specify the conditions under which experimental operations are performed. We annotate the activity for which the conditions apply as the MeasuredProperty and experiment participants which are worked on under these conditions as the MeasuredEntity (Table \ref{tab:data-annot-experimental-conditions}). 
    \begin{table}[!htbp]
    \centering
    \resizebox{\textwidth}{!}{
    \begin{tabular}{@{}llll@{}}
    \toprule
    \textbf{Sentence} &
      \multicolumn{3}{l}{\begin{tabular}[c]{@{}l@{}}
      Cleaned sponge and  diatom opal was dissolved via  wet alkaline \\ digestion at 100 °C for 40 min.
      \end{tabular}} \\ \midrule
    \textbf{}                                                                                   & \textbf{Quantity} & \textbf{MeasuredProperty} & \textbf{MeasuredEntity} \\ \midrule
    \multicolumn{1}{l|}{\textbf{\begin{tabular}[c]{@{}l@{}}Our \\ guideline\end{tabular}}} &
      100 °C &
      \begin{tabular}[c]{@{}l@{}}wet alkaline \\ digestion\end{tabular} &
      \begin{tabular}[c]{@{}l@{}}Cleaned sponge and \\ diatom opal\end{tabular} \\ \midrule
    \multicolumn{1}{l|}{\textbf{\begin{tabular}[c]{@{}l@{}}MeasEval \\ guideline\end{tabular}}} & 100 °C            &                           & wet alkaline digestion  \\ \bottomrule
    \end{tabular}
    }
    \caption{Example for the annotation of experimental conditions. MeasEval annotations taken from \citet{harper-etal-2021-semeval}'s corpus.}
    \label{tab:data-annot-experimental-conditions}
    \end{table}
    \item[Transformations] Experimental procedures often describe transformations of the MeasuredEntities before a measurable operation occurs. It is often not possible to pin-point one particular noun phrase that represents the entity to which the operation is being applied. Thus, we annotate all prior steps that are relevant for the operation as the MeasuredEntity. (Fig.  \ref{tab:data-annot-transformations}) 
    \begin{table}[!htbp]
    \centering
    \resizebox{\textwidth}{!}{
    \begin{tabular}{@{}llll@{}}
    \toprule
    \textbf{Sentence} &
      \multicolumn{3}{l}{\begin{tabular}[c]{@{}l@{}}To prepare C3N4-Pd composites, the as-prepared g-C3N4 was \\ added into 100 mL ethanol and was sonicated for 2 h to \\ obtain thin g-C3N4 nanosheets.\end{tabular}} \\ \midrule
    \textbf{}                                                                                   & \textbf{Quantity} & \textbf{MeasuredProperty} & \textbf{MeasuredEntity} \\ \midrule
    \multicolumn{1}{l|}{\textbf{\begin{tabular}[c]{@{}l@{}}Our \\ guideline\end{tabular}}} &
      2 h &
      \begin{tabular}[c]{@{}l@{}}sonicated \end{tabular} &
      \begin{tabular}[c]{@{}l@{}}as-prepared g-C3N4 was \\ added into 100 mL ethanol \end{tabular} \\ \midrule
    \multicolumn{1}{l|}{\textbf{\begin{tabular}[c]{@{}l@{}}MeasEval \\ guideline\end{tabular}}} & 2 h           &                           & sonicated  \\ \bottomrule
    \end{tabular}
    }
    \caption{MSP example for the annotation of experimentally transformed MeasuredEntities}
    \label{tab:data-annot-transformations}
    \end{table}
\end{description}
Although these guidelines deviate from the original MeasEval annotation guidelines, we believe that these rules are appropriate exceptions to accommodate the nature of experimental procedures, as these rules promote more information regarding measurements to be extracted. This goes in the direction of the "multiple hypothesis hypothesis" proposed by the authors of the MeasEval task, wherein they postulate that different interpretations of contextual information can be useful in different downstream applications \cite{harper-etal-2021-semeval}.

\section{Inter-annotator-agreement study}\label{appendix-iaa}
We conducted an IAA study for the re-annotation of the MSP dataset which spanned five rounds. For the annotation procedure we used the annotation tool prodigy \footnote{https://prodi.gy/}. After each round, the IAA was analyzed both through comparing the agreement score and the annotations themselves. The final annotation was chosen by selecting the annotation on which most annotators agreed. When there was no agreement, a discussion with all annotators decided either on the solution that adhered most closely to the existing guidelines or an amendment to the guidelines. 
As agreement measures, we calculate Krippendorff's Alpha coefficient \citep{Krippendorff.2004}. To ensure comparability, we follow the same implementation steps as the MeasEval authors and calculate the disagreement on the char-level using the python package simpledorff \footnote{https://github.com/LightTag/simpledorff}. Under that assumption, each character in an annotation sample is treated as a "markable" entity with its own label. 
\begin{figure}[!htbp]
\centering
\includegraphics[width=0.8\textwidth]{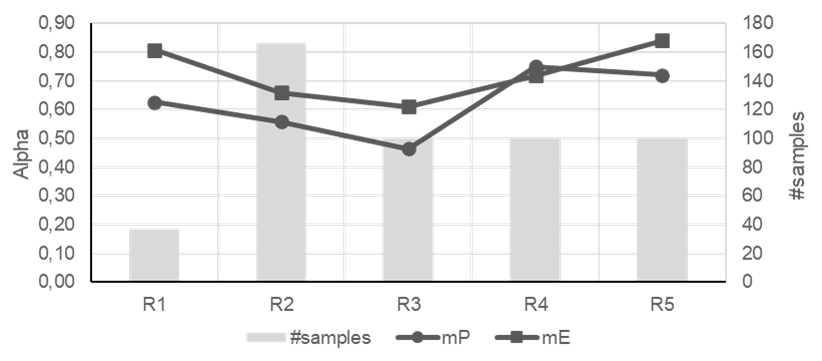}
\caption{Krippendorff's Alpha over five annotation rounds on mapping MSP to the MeasEval data model.}
\label{fig:data-krippendorff}
\end{figure}
Figure \ref{fig:data-krippendorff} shows the development of agreement for the annotation of MeasuredEntity and MeasuredProperty over the five rounds
The scores for the Unit and Value entities agreement were always near 1.0 and thus excluded from the analysis. The dip in agreement in round three was mainly due to a conflicting understanding of the supposed span length of MeasuredProperties. Having resolved this conflict, a "substantial" \citep{Viera.2005} agreement $>$ 0.67 could be achieved in round four and reproduced in round five. 

Although there were a number of ambiguous cases, the structure and content of experimental descriptions is mostly simple and formulaic. This is reflected in the moderate to high IAA scores compared to the MeasEval IAA, where the scores for both ME (0.55) and MP (0.64) are lower. 

\section{Corpus Overview} \label{appendix-corpus-overview}
Table \ref{tab:data-corpus-overview} details the main characteristics of each dataset.

\begin{table}[!htbp]
\resizebox{\textwidth}{!}{%
\begin{tabular}{@{}llccc@{}}
\toprule
 &
   &
  \textbf{MeasEval} &
  \textbf{MSP} &
  \textbf{Battery Materials} \\ \midrule
\multicolumn{2}{c}{\# Sentences/claims} &
  1250, 415, 724 &
  860, 89, 128 &
  194, 86, 72 \\\hline
  \multicolumn{2}{c}{\#Unigrams} &
  38,897 &
  17,062 &
  16,788 \\
  \multicolumn{2}{c}{\#Unique unigrams} &
  9,029 &
  4,024 &
  1,382\\
   \multicolumn{2}{c}{Ratio} &
  23\% &
  24\% &
  8\%\\\hline
\multirow{2}{*}{Quantity} &
  \begin{tabular}[c]{@{}l@{}}Total ents\end{tabular} &
  882, 281, 499 &
  1671, 195, 201 &
  278, 118, 102 \\
 &
  \begin{tabular}[c]{@{}l@{}}Unique/total ents\end{tabular} &
  0.83, 0.89, 0.81 &
  0.5, 0.72, 0.71 &
  0.62, 0.69, 0.78 \\\hline\hline
\multirow{3}{*}{\begin{tabular}[c]{@{}l@{}}Measured\\ Entity\end{tabular}} &
  \begin{tabular}[c]{@{}l@{}}Total ents\end{tabular} &
  875, 273, 499 &
  1669, 193, 199 &
  278, 118, 102 \\
 &
  \begin{tabular}[c]{@{}l@{}}Unique/total ents\end{tabular} &
  0.7, 0.6, 0.7 &
  0.42, 0.61, 0.67 &
  0.36, 0.36, 0.55 \\
 &
  Example ents &
  \textit{\begin{tabular}[c]{@{}c@{}}'cells', 'electrons', \\ 'samples', 'soil'\end{tabular}} &
  \textit{\begin{tabular}[c]{@{}c@{}}mixture', 'solution', \\ 'reaction', 'V2O5'\end{tabular}} &
  \textit{\begin{tabular}[c]{@{}c@{}}'secondary particles', \\ 'lithium metal oxide powder', \\ 'precursor'\end{tabular}} \\\hline\hline
\multirow{3}{*}{\begin{tabular}[c]{@{}l@{}}Measured\\ Property\end{tabular}} &
  \begin{tabular}[c]{@{}l@{}}Total entities \end{tabular} &
  563, 179, 330 &
  1379, 145, 157 &
  263, 118, 99 \\
 &
  \begin{tabular}[c]{@{}l@{}}Unique/total ents\end{tabular} &
  0.7, 0.61, 0.71 &
  0.28, 0.41, 0.48 &
  0.2, 0.34, 0.31 \\
 &
  Example ents &
  \textit{\begin{tabular}[c]{@{}c@{}}'n', 'depth', 'p', \\ 'odds ratios',  'ratio',\end{tabular}} &
  \textit{\begin{tabular}[c]{@{}c@{}}dissolved', 'dried', \\ 'calcined', 'heated'\end{tabular}} &
  \textit{\begin{tabular}[c]{@{}c@{}}particle size distribution', \\ 'tap density', 'sodium level', \\ 'average particle size',\end{tabular}} \\ \bottomrule
\end{tabular}}
\caption{Main characteristics of the datasets by data split (train, val, test)}
\label{tab:data-corpus-overview}
\end{table}

\section{Domain similarity}
Following the approach of \citet{Gururangan.2020}, we investigate the domain similarity of our datasets by studying their vocabulary overlap. The vocabulary overlap is based on the ratio of shared unigrams which we gather by tokenizing the texts with scispacy \footnote{https://allenai.github.io/scispacy/; en\_core\_sci\_lg model}. 

\begin{figure}[!htbp]
\centering
\includegraphics[width=\textwidth]{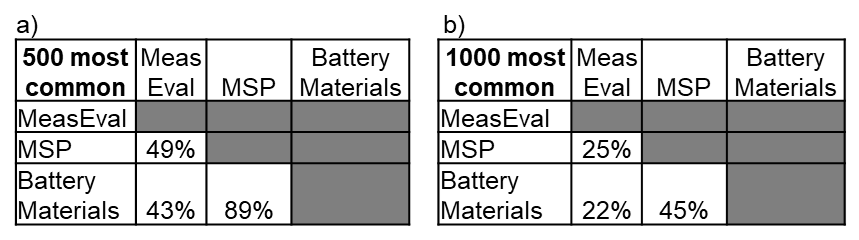}
\caption[Vocabulary overlap between datasets]{Vocabulary overlap between datasets: a) Overlap over 500 most common unigrams, b) Overlap over 1000 most common unigrams}
\label{fig:data-vocabuary-overlap}
\end{figure}

The matrices in Figure \ref{fig:data-vocabuary-overlap} show the resulting vocabulary overlap of the three datasets. They highlight the similarity between the MSP and BM dataset, which is especially pronounced in the comparison of the 500 most common unigrams with an overlap of 89\%. All in all, we assume that the MSP and BM corpus share the most similarity, followed by MeasEval and MSP and MeasEval and BM.

\section{Implementation Details}\label{appendix-implementation-details}
Below we lay out our implementation details for pre-training, and fine-tuning of the model setup. 
\begin{table}[!htbp]
\resizebox{\textwidth}{!}{
\begin{tabular}{@{}ll@{}}
\toprule
Computing infrastructure  & Ubuntu 18.04.6 LTS (GNU/Linux   5.4.0) \\ 
CUDA Version & 11.6 \\
GPU Type & Tesla V100-SXM2-32GB  \\
Available GPUs & 8 \\
Python version & 3.8                                    \\ \bottomrule
\end{tabular}
}
\caption{Environment details}
\label{tab:infrastructure}
\end{table}

Table \ref{tab:infrastructure} describes our computational infrastructure. We intermediately pre-train and fine-tune our base-models BERT-base-uncased (bert-base-uncased) and SciBERT-uncased (allenai/scibert\_scivocab\_uncased) which both have a 12 hidden layers with a hidden size of 768.

\paragraph{Adaptive Pre-training} Full intermediate pre-training was carried out using the masked language modeling script provided by the Huggingface Transformers Library \footnote{\url{https://github.com/huggingface/transformers/blob/main/examples/pytorch/language-modeling/run_mlm.py}}. For adapter-based pre-training we use AdapterHub \citep{Pfeiffer.2020} as well as their script for masked language modeling \footnote{\url{https://github.com/adapter-hub/adapter-transformers/blob/master/examples/pytorch/language-modeling/run_mlm.py}}. The hyperparameters are given in Table \ref{tab:adapter-hyperparameters}. Adapter config and reduction factor are adapter pre-training exclusive parameters. Except for the parameters in the table we use the default values provided by the script. For adaptive pre-training we did performed no systematic hyperparameter search, so there might be more optimal parameter settings. 

\begin{table}[!ht]
\resizebox{\textwidth}{!}{
\begin{tabular}{@{}ll@{}}
\toprule
Implementation   framework & huggingface/ AdapterHub  run-mlm.py script\\ 
optimizer                  & Adam                    \\
adam betas & 0.9, 0.98 \\
adam epsilon & 1e-06 \\
adapter config & pfeiffer+inv \\
reduction factor & 12 \\
learning rate              & 0.0001                  \\
bs                         & 64                      \\
lr scheduler type          & linear                  \\
lr scheduler warmup steps  & 100                     \\
num epochs                 & 40                      \\
evaluation strategy        & epoch  \\
seed                       & 42                      \\ \bottomrule
\end{tabular}
}
\caption{Pre-training hyperparameters for (adapter) TAPT. Pre-training was implemented based on the run-mlm.py script provided by huggingface / AdapterHub.}
\label{tab:adapter-hyperparameters}
\end{table}

\paragraph{Hyperparameter search for fine-tuning}
Hyperparameter tuning was performed using the ray.tune optimization framework for scalable hyperparameter tuning \footnote{\url{https://docs.ray.io/en/latest/tune/index.html}}. The tuning details are shown in Tables \ref{tab:fine-tuning-hyperparameters}. 
The training and validation loops are implemented with pytorch-lightning \footnote{\url{https://pytorch-lightning.readthedocs.io/en/stable/}}, a research framework built on pytorch \footnote{\url{https://pytorch.org/docs/stable/index.html}}. We train the models for Task 1 and Task 2 independently from each other, meaning that we train and tune our Task 2 models based on gold Quantities instead of prediction outputs from a Task 1 model. This is done by pre-enriching the Task 2 training sequences with special tokens ([Q] and [\/Q]) based on gold Quantity spans which simulates a perfect Task 1 performance. For future work, it might be also interesting to train and tune on the end-to-end pipeline. 
We optimize the models based the development strict F1 score, which is calculated by comparing predicted and gold BIO-tag sequences. We find that most models have their best parameter setting at a learning rate of 1e-05 or 5e-5 and a batch size of 16 or 32. Only the adapter-based models benefit from larger training rates of 1e-4 or 2e-4.
\begin{table}[!htbp]
\resizebox{\textwidth}{!}{
\begin{tabular}{@{}ll@{}}
\toprule
Implementation   framework    & pytorch-lightning + ray tune   \\ 
scheduler                     & ASHA scheduler                  \\
optimizer                     & Adam                          \\
max length                    & 512                           \\
max epochs                    & 15                            \\
patience                      & 5                             \\
gradient clipping             & max norm 1.0                  \\
lr                            & {[}1e-05, 5e-5, 1e-4, 2e-4{]} \\
bs                            & {[}8, 16, 32, 64{]}           \\
weight decay                  & 0.01                          \\
stochastic weight averaging & yes                           \\
seed                          & 1                             \\ \bottomrule
\end{tabular}
}
\caption{Fine-tuning hyperparameters for Task 1 and Task 2}
\label{tab:fine-tuning-hyperparameters}
\end{table}

\section{Test Results Tables}\label{appendix-test-results}
Table \ref{tab:results-t1-test-strictf1} shows the complete results table on the test data of Task 1. Table \ref{tab:results-t2-test-strictf1} shows the complete results table on the test data of Task 2. All scores refer to the strict F1, not overlap F1.
\begin{table*}[!ht]
\centering
\resizebox{\textwidth}{!}{%
\begin{tabular}{lcllcccc}
\toprule
 & \multicolumn{1}{l}{} &  &  & \multicolumn{4}{c}{Target Domain} \\ \cmidrule{5-8}
\begin{tabular}[c]{@{}l@{}}Training \\ Mode\end{tabular} & \multicolumn{1}{l}{\begin{tabular}[c]{@{}l@{}}Source \\ Domain\end{tabular}} & Model & \begin{tabular}[c]{@{}l@{}}PT \\ Setup\end{tabular} &\begin{tabular}[c]{@{}l@{}}Meas \\ Eval\end{tabular} & MSP & BM & Overall \\ \hline \addlinespace
\multirow{18}{*}{\begin{tabular}[c]{@{}l@{}}Single-\\ Source\end{tabular}} & \multirow{6}{*}{MeasEval} & \multirow{3}{*}{BERT} & No PT & 0.777 & 0.697 & 0.385 & 0.671 \\
 &  &  & Full PT & 0.759 & 0.849 & \textit{0.409} & 0.702 \\
 &  &  & Adapter PT & 0.749 & 0.840 & 0.369 & 0.688 \\\cmidrule{3-8}
 &  & \multirow{3}{*}{SciBERT} & No PT & \textit{\textbf{0.786}} & 0.897 & 0.396 & \textit{0.721} \\
 &  &  & Full PT & \textit{\textbf{0.786}} & \textit{0.913} & 0.378 & 0.719 \\
 &  &  & Adapter PT & 0.783 & 0.889 & 0.377 & 0.715 \\\cmidrule{2-8}
 & \multirow{6}{*}{MSP} & \multirow{3}{*}{BERT} & No PT & 0.627 & 0.924 & 0.369 & 0.632 \\
 &  &  & Full PT & 0.637 & 0.923 & 0.344 & 0.634 \\
 &  &  & Adapter PT & 0.558 & 0.893 & 0.424 & 0.607 \\\cmidrule{3-8}
 &  & \multirow{3}{*}{SciBERT} & No PT & 0.653 & 0.908 & \textit{0.455} & \textit{0.667} \\
 &  &  & Full PT & 0.656 & 0.913 & 0.439 & 0.667 \\
 &  &  & Adapter PT & \textit{0.659} & \textit{0.935} & 0.418 & 0.665 \\\cmidrule{2-8}
 & \multirow{6}{*}{BM} & \multirow{3}{*}{BERT} & No PT & 0.256 & 0.136 & 0.500 & 0.290 \\
 &  &  & Full PT & 0.255 & 0.182 & 0.451 & 0.285 \\
 &  &  & Adapter PT & 0.271 & 0.282 & 0.442 & 0.315 \\\cmidrule{3-8}
 &  & \multirow{3}{*}{SciBERT} & No PT & \textit{0.354} & \textit{0.339} & \textit{\textbf{0.521}} & \textit{0.386} \\
 &  &  & Full PT & 0.324 & 0.301 & 0.505 & 0.357 \\
 &  &  & Adapter PT & 0.208 & 0.049 & 0.387 & 0.222 \\ \midrule
\multirow{6}{*}{\begin{tabular}[c]{@{}l@{}}Duo-\\ Source\end{tabular}} & \multirow{6}{*}{\begin{tabular}[c]{@{}c@{}}MSP+\\ MeasEval\end{tabular}} & \multirow{3}{*}{BERT} & No PT & 0.744 & 0.915 & 0.402 & 0.710 \\
 &  &  & Full PT & 0.761 & 0.893 & 0.448 & 0.726 \\
 &  &  & Adapter PT & 0.763 & \textit{\textbf{0.938}} & 0.422 & 0.732 \\\cmidrule{3-8}
 &  & \multirow{3}{*}{SciBERT} & No PT & \textit{0.778} & 0.908 & \textit{0.457} & \textit{\textbf{0.739}} \\
 &  &  & Full PT & 0.776 & 0.925 & 0.398 & 0.721 \\
 &  &  & Adapter PT & 0.767 & 0.935 & 0.417 & 0.727 \\ \bottomrule
\end{tabular}%
}
\caption[F1 scores of Task 1: Quantity Extraction]{Test F1 scores of Task 1: Quantity Extraction. \textbf{Bold} scores indicate the highest score across an entire target domain. \textit{Italic} scores indicate the highest score within one source domain.}
\label{tab:results-t1-test-strictf1}
\end{table*} 
\begin{table*}[!ht]
\centering
\resizebox{\textwidth}{!}{%
\begin{tabular}{@{}c|c|c|l|cccc|cccc|cccc|cccc@{}}
\toprule
\multicolumn{1}{l}{} &
  \multicolumn{1}{l}{} &
  \multicolumn{1}{l}{} &
   &
  \multicolumn{16}{c}{Target Domain} \\ 
\multicolumn{1}{l}{} &
  \multicolumn{1}{l}{} &
  \multicolumn{1}{l}{} &
   &
  \multicolumn{4}{c}{MeasEval} &
  \multicolumn{4}{c}{MSP} &
  \multicolumn{4}{c}{BM} &
  \multicolumn{4}{c}{Overall} \\\cmidrule{5-20}
\multicolumn{1}{l}{\begin{tabular}[c]{@{}l@{}}Training \\ Mode\end{tabular}} &
  \multicolumn{1}{l}{\begin{tabular}[c]{@{}l@{}}Source \\ Domain\end{tabular}} &
  \multicolumn{1}{l}{Model} &
  PT Setup &
  O &
  U &
  ME &
  MP &
  O &
  U &
  ME &
  MP &
  O &
  U &
  ME &
  MP &
  O &
  U &
  ME &
  MP \\ \toprule
\multirow{18}{*}{\begin{tabular}[c]{@{}c@{}}Single-\\ Source\end{tabular}} &
  \multirow{6}{*}{\begin{tabular}[c]{@{}c@{}}Meas\\ Eval\end{tabular}} &
  \multirow{3}{*}{BERT} &
  No PT &
  0.602 &
  0.949 &
  0.436 &
  0.436 &
  0.663 &
  0.984 &
  0.463 &
  0.502 &
  0.627 &
  \textit{0.957} &
  0.470 &
  0.563 &
  0.621 &
  0.960 &
  0.448 &
  0.473 \\
 &
   &
   &
  Full PT &
  0.591 &
  0.963 &
  0.416 &
  0.411 &
  0.624 &
  0.978 &
  0.436 &
  0.434 &
  0.604 &
  0.917 &
  0.436 &
  0.574 &
  0.602 &
  0.963 &
  0.424 &
  0.446 \\
 &
   &
   &
  Adapter PT &
  0.540 &
  0.926 &
  0.366 &
  0.385 &
  0.598 &
  0.984 &
  0.393 &
  0.400 &
  0.561 &
  0.829 &
  0.480 &
  0.469 &
  0.557 &
  0.932 &
  0.388 &
  0.403 \\ \cmidrule{3-20}
 &
   &
  \multirow{3}{*}{SciBERT} &
  No PT &
  \textit{0.638} &
  \textit{\textbf{0.970}} &
  \textit{0.494} &
  \textit{0.460} &
  \textit{0.686} &
  \textit{0.995} &
  \textit{0.494} &
  \textit{0.563} &
  0.676 &
  0.929 &
  0.545 &
  0.667 &
  \textit{0.655} &
  \textit{0.972} &
  \textit{0.501} &
  \textit{0.522} \\
 &
   &
   &
  Full PT &
  0.628 &
  0.958 &
  0.481 &
  0.450 &
  0.648 &
  0.989 &
  0.468 &
  0.468 &
  \textit{\textbf{0.718}} &
  0.900 &
  \textit{0.584} &
  \textit{0.726} &
  0.646 &
  0.961 &
  0.491 &
  0.508 \\
 &
   &
   &
  Adapter PT &
  0.574 &
  0.951 &
  0.424 &
  0.369 &
  0.625 &
  0.973 &
  0.405 &
  0.506 &
  0.604 &
  0.905 &
  0.519 &
  0.502 &
  0.591 &
  0.952 &
  0.431 &
  0.425 \\\cmidrule{2-20}
 &
  \multirow{6}{*}{MSP} &
  \multirow{3}{*}{BERT} &
  No PT &
  0.547 &
  0.953 &
  \textit{0.385} &
  0.316 &
  \textit{\textbf{0.750}} &
  \textit{\textbf{1.000}} &
  \textit{\textbf{0.589}} &
  0.673 &
  0.627 &
  \textit{\textbf{0.971}} &
  0.505 &
  0.514 &
  0.612 &
  \textit{0.968} &
  0.452 &
  0.445 \\
 &
   &
   &
  Full PT &
  0.530 &
  0.943 &
  0.361 &
  0.316 &
  \textit{\textbf{0.751}} &
  0.997 &
  0.571 &
  0.697 &
  0.661 &
  0.922 &
  0.510 &
  \textit{0.633} &
  0.606 &
  0.956 &
  0.437 &
  0.461 \\
 &
   &
   &
  Adapter PT &
  0.509 &
  0.944 &
  0.332 &
  0.291 &
  0.730 &
  0.995 &
  0.538 &
  0.677 &
  0.536 &
  0.829 &
  0.395 &
  0.498 &
  0.570 &
  0.946 &
  0.392 &
  0.421 \\\cmidrule{3-20}
 &
   &
  \multirow{3}{*}{SciBERT} &
  No PT &
  \textit{0.567} &
  0.954 &
  0.372 &
  0.371 &
  0.744 &
  \textit{\textbf{1.000}} &
  0.556 &
  0.687 &
  \textit{0.690} &
  0.922 &
  \textit{\textbf{0.622}} &
  0.615 &
  \textit{0.633} &
  0.964 &
  \textit{0.456} &
  \textit{0.502} \\
 &
   &
   &
  Full PT &
  0.561 &
  \textit{0.961} &
  0.368 &
  \textit{0.379} &
  0.748 &
  \textit{\textbf{1.000}} &
  \textit{\textbf{0.580}} &
  0.684 &
  0.668 &
  0.901 &
  0.558 &
  0.625 &
  0.626 &
  0.965 &
  0.450 &
  0.500 \\
 &
   &
   &
  Adapter PT &
  0.531 &
  0.948 &
  0.310 &
  0.365 &
  0.737 &
  0.995 &
  0.532 &
  \textit{\textbf{0.711}} &
  0.608 &
  0.863 &
  0.479 &
  0.588 &
  0.597 &
  0.952 &
  0.393 &
  0.491 \\\cmidrule{2-20}
 &
  \multirow{6}{*}{BM} &
  \multirow{3}{*}{BERT} &
  No PT &
  0.353 &
  \textit{0.898} &
  0.094 &
  0.141 &
  0.382 &
  0.755 &
  0.157 &
  0.102 &
  0.575 &
  \textit{\textbf{0.971}} &
  0.229 &
  0.628 &
  0.389 &
  \textit{0.865} &
  0.125 &
  0.216 \\
 &
   &
   &
  Full PT &
  0.352 &
  0.801 &
  0.130 &
  0.202 &
  0.379 &
  0.826 &
  0.124 &
  0.180 &
  0.618 &
  0.971 &
  0.294 &
  0.694 &
  0.396 &
  0.828 &
  0.148 &
  0.297 \\
 &
   &
   &
  Adapter PT &
  0.305 &
  0.782 &
  0.076 &
  0.151 &
  \textit{0.410} &
  \textit{0.828} &
  0.189 &
  0.121 &
  0.524 &
  0.906 &
  0.187 &
  0.634 &
  0.363 &
  0.810 &
  0.119 &
  0.238 \\\cmidrule{3-20}
 &
   &
  \multirow{3}{*}{SciBERT} &
  No PT &
  0.366 &
  0.759 &
  \textit{0.195} &
  \textit{0.264} &
  0.356 &
  0.646 &
  \textit{0.251} &
  \textit{0.195} &
  \textit{0.650} &
  0.929 &
  \textit{0.398} &
  0.738 &
  0.405 &
  0.750 &
  \textit{0.235} &
  \textit{0.355} \\
 &
   &
   &
  Full PT &
  \textit{0.375} &
  0.779 &
  0.166 &
  0.220 &
  0.344 &
  0.634 &
  0.214 &
  0.185 &
  0.648 &
  0.900 &
  0.387 &
  \textit{\textbf{0.757}} &
  \textit{0.409} &
  0.755 &
  0.210 &
  0.329 \\
 &
   &
   &
  Adapter PT &
  0.249 &
  0.548 &
  0.108 &
  0.153 &
  0.218 &
  0.417 &
  0.135 &
  0.099 &
  0.570 &
  0.882 &
  0.292 &
  0.680 &
  0.296 &
  0.559 &
  0.143 &
  0.263 \\\bottomrule\bottomrule
\multirow{12}{*}{\begin{tabular}[c]{@{}c@{}}Duo-\\ Source\end{tabular}} &
  \multirow{4}{*}{\begin{tabular}[c]{@{}c@{}}BM + \\ Meas\\ Eval\end{tabular}} &
  \multirow{2}{*}{BERT} &
  No PT &
  0.558 &
  0.959 &
  0.380 &
  0.389 &
  0.618 &
  0.992 &
  0.435 &
  0.428 &
  0.633 &
  0.950 &
  0.439 &
  0.631 &
  0.584 &
  0.968 &
  0.401 &
  0.441 \\
 &
   &
   &
  Full PT &
  0.604 &
  0.957 &
  0.440 &
  0.431 &
  0.627 &
  0.989 &
  0.421 &
  0.440 &
  0.658 &
  \textit{\textbf{0.971}} &
  \textit{0.466} &
  0.650 &
  0.617 &
  0.967 &
  0.439 &
  0.474 \\ \cmidrule{3-20}
 &
   &
  \multirow{2}{*}{SciBERT} &
  No PT &
  0.639 &
  \textit{\textbf{0.968}} &
  0.489 &
  0.471 &
  0.660 &
  \textit{0.995} &
  0.450 &
  \textit{0.532} &
  \textit{0.678} &
  0.914 &
  0.443 &
  \textit{\textbf{0.769}} &
  0.650 &
  \textit{0.970} &
  0.472 &
  \textit{0.540} \\
 &
   &
   &
  Full PT &
  \textit{\textbf{0.647}} &
  0.962 &
  \textit{\textbf{0.503}} &
  \textit{\textbf{0.490}} &
  \textit{0.662} &
  0.989 &
  \textit{0.464} &
  0.522 &
  0.653 &
  0.914 &
  0.423 &
  0.709 &
  \textit{0.652} &
  0.965 &
  \textit{0.482} &
  0.537 \\ \cmidrule{2-20}
 &
  \multirow{4}{*}{\begin{tabular}[c]{@{}c@{}}BM +\\ MSP\end{tabular}} &
  \multirow{2}{*}{BERT} &
  No PT &
  0.560 &
  \textit{0.965} &
  0.394 &
  0.340 &
  0.746 &
  0.992 &
  0.574 &
  0.694 &
  0.675 &
  \textit{0.957} &
  0.500 &
  0.667 &
  0.626 &
  \textit{0.972} &
  \textit{0.455} &
  0.489 \\
 &
   &
   &
  Full PT &
  0.536 &
  0.958 &
  0.340 &
  0.343 &
  0.736 &
  0.989 &
  0.545 &
  \textit{0.696} &
  \textit{0.695} &
  \textit{0.957} &
  \textit{0.534} &
  \textit{0.689} &
  0.610 &
  0.967 &
  0.417 &
  0.496 \\ \cmidrule{3-20}
 &
   &
  \multirow{2}{*}{SciBERT} &
  No PT &
  0.559 &
  0.961 &
  0.373 &
  \textit{0.379} &
  \textit{\textbf{0.750}} &
  \textit{\textbf{1.000}} &
  \textit{0.579} &
  0.688 &
  0.641 &
  0.914 &
  0.434 &
  0.676 &
  0.621 &
  0.967 &
  0.435 &
  \textit{0.506} \\
 &
   &
   &
  Full PT &
  \textit{0.570} &
  0.956 &
  \textit{0.404} &
  0.367 &
  0.738 &
  \textit{\textbf{1.000}} &
  0.542 &
  0.695 &
  0.650 &
  0.908 &
  0.459 &
  0.667 &
  \textit{0.626} &
  0.963 &
  0.448 &
  0.502 \\ \cmidrule{2-20}
 &
  \multirow{4}{*}{\begin{tabular}[c]{@{}c@{}}MSP +\\ Meas\\ Eval\end{tabular}} &
  \multirow{2}{*}{BERT} &
  No PT &
  0.599 &
  0.962 &
  0.437 &
  0.420 &
  0.745 &
  0.995 &
  0.551 &
  \textit{\textbf{0.716}} &
  0.669 &
  0.929 &
  0.511 &
  0.660 &
  0.648 &
  0.968 &
  0.477 &
  0.538 \\
 &
   &
   &
  Full PT &
  0.611 &
  0.966 &
  0.471 &
  0.396 &
  0.736 &
  0.997 &
  0.560 &
  0.662 &
  0.665 &
  \textit{0.957} &
  0.585 &
  0.554 &
  0.651 &
  \textit{\textbf{0.974}} &
  0.508 &
  0.496 \\ \cmidrule{3-20}
 &
   &
  \multirow{2}{*}{SciBERT} &
  No PT &
  \textit{\textbf{0.651}} &
  0.962 &
  \textit{\textbf{0.499}} &
  \textit{\textbf{0.510}} &
  \textit{0.749} &
  \textit{\textbf{1.000}} &
  0.568 &
  0.704 &
  \textit{\textbf{0.721}} &
  0.929 &
  \textit{\textbf{0.622}} &
  0.682 &
  \textit{\textbf{0.687}} &
  0.969 &
  \textit{\textbf{0.534}} &
  \textit{\textbf{0.589}} \\
 &
   &
   &
  Full PT &
  0.622 &
  \textit{\textbf{0.972}} &
  0.478 &
  0.446 &
  0.745 &
  \textit{\textbf{1.000}} &
  \textit{0.570} &
  0.687 &
  \textit{\textbf{0.746}} &
  0.930 &
  \textit{\textbf{0.652}} &
  \textit{0.721} &
  0.671 &
  \textit{\textbf{0.975}} &
  \textit{\textbf{0.523}} &
  0.557 \\\bottomrule\bottomrule
\multirow{6}{*}{\begin{tabular}[c]{@{}c@{}}All \\ Sources\end{tabular}} &
  \multirow{6}{*}{\begin{tabular}[c]{@{}c@{}}MSP+ \\ Meas\\ Eval + \\ BM\end{tabular}} &
  \multirow{3}{*}{BERT} &
  No PT &
  0.610 &
  0.962 &
  0.464 &
  0.410 &
  0.730 &
  \textit{\textbf{1.000}} &
  0.560 &
  0.629 &
  0.631 &
  \textit{\textbf{0.986}} &
  0.384 &
  0.634 &
  0.645 &
  \textit{\textbf{0.975}} &
  0.479 &
  0.506 \\
 &
   &
   &
  Full PT &
  0.609 &
  0.960 &
  0.445 &
  0.444 &
  0.747 &
  0.989 &
  \textit{\textbf{0.588}} &
  0.669 &
  0.595 &
  \textit{\textbf{0.971}} &
  0.276 &
  0.657 &
  0.644 &
  0.969 &
  0.460 &
  0.536 \\
 &
   &
   &
  Adapter PT &
  0.556 &
  0.959 &
  0.370 &
  0.375 &
  0.705 &
  0.995 &
  0.520 &
  0.627 &
  0.618 &
  0.957 &
  0.377 &
  0.634 &
  0.604 &
  0.969 &
  0.411 &
  0.484 \\
 &
   &
  \multirow{3}{*}{SciBERT} &
  No PT &
  0.634 &
  \textit{0.965} &
  0.482 &
  0.476 &
  \textit{0.748} &
  \textit{\textbf{1.000}} &
  0.562 &
  0.700 &
  0.698 &
  0.929 &
  \textit{0.546} &
  0.692 &
  \textit{\textbf{0.673}} &
  0.971 &
  0.512 &
  \textit{\textbf{0.569}} \\
 &
   &
   &
  Full PT &
  \textit{\textbf{0.654}} &
  0.963 &
  \textit{\textbf{0.515}} &
  \textit{\textbf{0.499}} &
  0.741 &
  \textit{\textbf{1.000}} &
  0.556 &
  0.691 &
  \textit{0.702} &
  0.943 &
  0.500 &
  \textit{\textbf{0.756}} &
  \textit{\textbf{0.684}} &
  0.972 &
  \textit{\textbf{0.524}} &
  \textit{\textbf{0.593}} \\
 &
   &
   &
  Adapter PT &
  0.575 &
  0.952 &
  0.413 &
  0.398 &
  0.737 &
  0.986 &
  0.547 &
  \textit{\textbf{0.719}} &
  0.667 &
  0.900 &
  0.466 &
  0.725 &
  0.630 &
  0.956 &
  0.454 &
  0.535 \\ \bottomrule
\end{tabular}%
}
\caption[F1 scores of Task 2: Context extraction]{Test F1 scores of Task 2: Context extraction. \textbf{Bold} scores indicate the highest score across an entire target domain. \textit{Italic} scores indicate the highest score within one source domain. O = Overall, U = Unit, ME = MeasuredEntity, MP = MeasuredProperty.}
\label{tab:results-t2-test-strictf1}
\end{table*}

\section{Development Result Tables}\label{appendix-development-results}
Tables \ref{tab:t1_results_val}, \ref{tab:dev_t2}, and \ref{tab:dev_e2e} show the Task 1, Task 2 and end-to-end results on the development portion of the corpus. All scores refer to the strict F1, not overlap F1.
\begin{table*}[!ht]
\centering
\resizebox{\textwidth}{!}{%
\begin{tabular}{@{}clllcccc@{}}
\toprule
\multicolumn{1}{l}{\begin{tabular}[c]{@{}l@{}}Training \\ mode\end{tabular}} & \begin{tabular}[c]{@{}l@{}}Source \\ domain\end{tabular} & Model & \multicolumn{1}{c}{PT Setup} & MeasEval & MSP & BM & Overall \\ \midrule
\multirow{18}{*}{\begin{tabular}[c]{@{}c@{}}Single-\\ Source\end{tabular}} & \multirow{6}{*}{\begin{tabular}[c]{@{}l@{}}Meas\\ Eval\end{tabular}} & \multirow{3}{*}{BERT} & No PT & 0.749 & 0.756 & 0.281 & 0.629 \\
 &  &  & Full PT & 0.757 & 0.856 & 0.285 & 0.661 \\
 &  &  & Adapter PT & 0.705 & 0.805 & 0.304 & 0.624 \\ \cmidrule{3-8}
 &  & \multirow{3}{*}{SciBERT} & No PT & \textbf{0.768} & 0.845 & 0.284 & 0.653 \\
 &  &  & Full PT & \textbf{0.774} & 0.855 & 0.302 & 0.663 \\
 &  &  & Adapter PT & 0.745 & 0.840 & 0.308 & 0.649 \\ \cmidrule{2-8}
 & \multirow{6}{*}{MSP} & \multirow{3}{*}{BERT} & No PT & 0.625 & 0.922 & 0.195 & 0.584 \\
 &  &  & Full PT & 0.584 & \textbf{0.927} & 0.187 & 0.581 \\
 &  &  & Adapter PT & 0.522 & 0.912 & 0.222 & 0.563 \\ \cmidrule{3-8}
 &  & \multirow{3}{*}{SciBERT} & No PT & 0.653 & \textbf{0.931} & 0.275 & 0.627 \\
 &  &  & Full PT & \multicolumn{1}{c}{0.643} & \multicolumn{1}{l}{0.924} & \multicolumn{1}{c}{0.268} & \multicolumn{1}{c}{0.621} \\
 &  &  & Adapter PT & \multicolumn{1}{c}{0.610} & \multicolumn{1}{l}{0.919} & \multicolumn{1}{c}{0.234} & \multicolumn{1}{l}{0.583} \\ \cmidrule{2-8}
 & \multirow{6}{*}{BM} & \multirow{3}{*}{BERT} & No PT & 0.296 & 0.103 & 0.612 & 0.329 \\
 &  &  & Full PT & 0.363 & 0.083 & \textbf{0.621} & 0.356 \\
 &  &  & Adapter PT & 0.293 & 0.171 & 0.586 & 0.337 \\\cmidrule{3-8}
 &  & \multirow{3}{*}{SciBERT} & No PT & 0.356 & 0.179 & \textbf{0.628} & 0.373 \\
 &  &  & Full PT & 0.361 & 0.228 & \textbf{0.669} & 0.399 \\
 &  &  & Adapter PT & 0.224 & 0.065 & 0.580 & 0.278 \\ \midrule
\multirow{6}{*}{\begin{tabular}[c]{@{}c@{}}Duo-\\ Source\end{tabular}} & \multicolumn{1}{c}{\multirow{6}{*}{\begin{tabular}[c]{@{}c@{}}MSP+\\ Meas\\ Eval\end{tabular}}} & \multirow{3}{*}{BERT} & No PT & 0.751 & 0.922 & 0.354 & \textbf{0.704} \\
 & \multicolumn{1}{c}{} &  & Full PT & 0.753 & 0.909 & 0.351 & \textbf{0.700} \\
 & \multicolumn{1}{c}{} &  & Adapter PT & 0.721 & 0.899 & 0.338 & 0.677 \\ \cmidrule{3-8}
 & \multicolumn{1}{c}{} & \multirow{3}{*}{SciBERT} & No PT & 0.756 & \textbf{0.937} & 0.324 & \textbf{0.687} \\
 & \multicolumn{1}{c}{} &  & Full PT & \textbf{0.762} & 0.909 & 0.313 & 0.681 \\
 & \multicolumn{1}{c}{} &  & Adapter PT & 0.756 & 0.897 & 0.320 & 0.676 \\ \bottomrule
\end{tabular}%
}
\caption[Development F1 scores of Task 1]{Development F1 scores of Task 1: Quantity Extraction. \textbf{Bold} scores indicate the highest score across an entire target domain. \textit{Italic} scores indicate the highest score within one source domain.}
\label{tab:t1_results_val}
\end{table*}

\begin{table*}[!ht]
\centering
\resizebox{\textwidth}{!}{%
\begin{tabular}{@{}clllcccccccccccccccc@{}}
\toprule
\multicolumn{1}{l}{} &
   &
   &
   &
  \multicolumn{4}{c}{MeasEval} &
  \multicolumn{4}{c}{MSP} &
  \multicolumn{4}{c}{BM} &
  \multicolumn{4}{c}{Overall} \\ \midrule
\multicolumn{1}{l}{\begin{tabular}[c]{@{}l@{}}Training \\ mode\end{tabular}} &
  \begin{tabular}[c]{@{}l@{}}Source \\ domain\end{tabular} &
  Model &
  \begin{tabular}[c]{@{}l@{}}PT \\ Setup\end{tabular} &
  All &
  U &
  ME &
  MP &
  All &
  U &
  ME &
  MP &
  All &
  U &
  ME &
  MP &
  All &
  U &
  ME &
  MP \\ \midrule
\multirow{18}{*}{\begin{tabular}[c]{@{}c@{}}Single-\\ Source\end{tabular}} &
  \multirow{6}{*}{\begin{tabular}[c]{@{}l@{}}Meas\\ Eval\end{tabular}} &
  \multirow{3}{*}{BERT} &
  No PT &
  0.558 &
  0.946 &
  0.370 &
  0.322 &
  0.614 &
  0.986 &
  0.356 &
  0.421 &
  0.451 &
  0.890 &
  0.133 &
  0.455 &
  0.553 &
  0.951 &
  0.315 &
  0.387 \\
 &
   &
   &
  Full PT &
  0.556 &
  0.946 &
  0.386 &
  0.287 &
  0.630 &
  0.992 &
  0.413 &
  0.456 &
  0.440 &
  0.838 &
  0.165 &
  0.436 &
  0.556 &
  0.943 &
  0.348 &
  0.381 \\
 &
   &
   &
  Adapter PT &
  0.508 &
  0.937 &
  0.322 &
  0.290 &
  0.576 &
  0.970 &
  0.361 &
  0.353 &
  0.380 &
  0.744 &
  0.136 &
  0.389 &
  0.503 &
  0.916 &
  0.295 &
  0.333 \\ \cmidrule{3-20} 
 &
   &
  \multirow{3}{*}{SciBERT} &
  No PT &
  0.582 &
  0.950 &
  \textbf{0.412} &
  0.357 &
  0.636 &
  0.984 &
  0.407 &
  0.504 &
  0.504 &
  0.849 &
  0.205 &
  0.596 &
  0.583 &
  0.944 &
  0.365 &
  0.469 \\
 &
   &
   &
  Full PT &
  0.579 &
  \textbf{0.954} &
  0.385 &
  \textbf{0.385} &
  0.667 &
  0.975 &
  0.468 &
  0.517 &
  0.547 &
  0.883 &
  0.239 &
  0.579 &
  0.602 &
  0.949 &
  0.384 &
  0.482 \\
 &
   &
   &
  Adapter PT &
  0.531 &
  \textbf{0.957} &
  0.340 &
  0.297 &
  0.634 &
  0.978 &
  0.408 &
  0.482 &
  0.451 &
  0.766 &
  0.192 &
  0.482 &
  0.548 &
  0.931 &
  0.333 &
  0.402 \\ \cmidrule{2-20} 
 &
  \multirow{6}{*}{MSP} &
  \multirow{3}{*}{BERT} &
  No PT &
  0.516 &
  0.943 &
  0.307 &
  0.302 &
  0.768 &
  0.992 &
  0.588 &
  \textbf{0.728} &
  0.522 &
  0.914 &
  0.216 &
  0.531 &
  0.604 &
  0.956 &
  0.384 &
  0.499 \\
 &
   &
   &
  Full PT &
  0.473 &
  0.946 &
  0.239 &
  0.253 &
  0.774 &
  \textbf{0.997} &
  0.612 &
  0.703 &
  0.508 &
  0.909 &
  0.189 &
  0.525 &
  0.581 &
  0.958 &
  0.355 &
  0.452 \\
 &
   &
   &
  Adapter PT &
  0.461 &
  0.915 &
  0.258 &
  0.240 &
  0.732 &
  0.989 &
  0.569 &
  0.644 &
  0.474 &
  0.836 &
  0.167 &
  0.542 &
  0.554 &
  0.928 &
  0.342 &
  0.440 \\ \cmidrule{3-20} 
 &
   &
  \multirow{3}{*}{SciBERT} &
  No PT &
  0.532 &
  0.948 &
  0.324 &
  0.291 &
  \textbf{0.796} &
  0.992 &
  \textbf{0.643} &
  \textbf{0.744} &
  0.477 &
  0.818 &
  0.167 &
  0.537 &
  0.610 &
  0.941 &
  0.395 &
  0.509 \\
 &
   &
   &
  Full PT &
  0.527 &
  0.950 &
  0.308 &
  0.329 &
  0.770 &
  0.986 &
  0.608 &
  0.724 &
  0.534 &
  0.807 &
  0.247 &
  0.624 &
  0.611 &
  0.938 &
  0.398 &
  0.531 \\
 &
   &
   &
  Adapter PT &
  0.475 &
  0.940 &
  0.227 &
  0.262 &
  0.744 &
  0.989 &
  0.549 &
  0.693 &
  0.456 &
  0.701 &
  0.175 &
  0.567 &
  0.561 &
  0.916 &
  0.323 &
  0.475 \\ \cmidrule{2-20} 
 &
  \multirow{6}{*}{BM} &
  \multirow{3}{*}{BERT} &
  No PT &
  0.372 &
  0.855 &
  0.076 &
  0.216 &
  0.398 &
  0.827 &
  0.094 &
  0.074 &
  0.651 &
  0.895 &
  0.385 &
  0.709 &
  0.439 &
  0.852 &
  0.145 &
  0.316 \\
 &
   &
   &
  Full PT &
  0.408 &
  0.859 &
  0.119 &
  0.262 &
  0.355 &
  0.784 &
  0.102 &
  0.098 &
  0.659 &
  \textbf{0.955} &
  \textbf{0.417} &
  0.672 &
  0.450 &
  0.852 &
  0.175 &
  0.354 \\
 &
   &
   &
  Adapter PT &
  0.347 &
  0.810 &
  0.052 &
  0.234 &
  0.385 &
  0.840 &
  0.114 &
  0.079 &
  0.595 &
  0.881 &
  0.337 &
  0.650 &
  0.415 &
  0.834 &
  0.134 &
  0.325 \\ \cmidrule{3-20} 
 &
   &
  \multirow{3}{*}{SciBERT} &
  No PT &
  0.377 &
  0.791 &
  0.149 &
  0.323 &
  0.268 &
  0.508 &
  0.188 &
  0.126 &
  0.659 &
  0.909 &
  \textbf{0.452} &
  0.688 &
  0.407 &
  0.729 &
  0.216 &
  0.396 \\
 &
   &
   &
  Full PT &
  0.398 &
  0.824 &
  0.112 &
  0.292 &
  0.304 &
  0.568 &
  0.205 &
  0.104 &
  0.665 &
  0.933 &
  0.370 &
  \textbf{0.763} &
  0.432 &
  0.764 &
  0.198 &
  0.403 \\
 &
   &
   &
  Adapter PT &
  0.300 &
  0.611 &
  0.119 &
  0.199 &
  0.236 &
  0.405 &
  0.176 &
  0.099 &
  0.618 &
  0.807 &
  0.389 &
  0.727 &
  0.365 &
  0.588 &
  0.203 &
  0.359 \\ \midrule
\multirow{12}{*}{\begin{tabular}[c]{@{}c@{}}Duo-\\ Source\end{tabular}} &
  \multirow{4}{*}{\begin{tabular}[c]{@{}l@{}}BM +\\ Meas\\ Eval\end{tabular}} &
  \multirow{2}{*}{BERT} &
  No PT &
  0.532 &
  0.937 &
  0.338 &
  0.354 &
  0.587 &
  0.984 &
  0.347 &
  0.439 &
  0.648 &
  0.939 &
  0.411 &
  0.686 &
  0.575 &
  0.954 &
  0.355 &
  0.470 \\
 &
   &
   &
  Full PT &
  0.556 &
  0.936 &
  0.369 &
  0.353 &
  0.627 &
  0.989 &
  0.410 &
  0.454 &
  0.650 &
  \textbf{0.961} &
  0.369 &
  0.721 &
  0.601 &
  0.960 &
  0.382 &
  0.491 \\ \cmidrule{3-20} 
 &
   &
  \multirow{2}{*}{SciBERT} &
  No PT &
  0.579 &
  \textbf{0.952} &
  0.400 &
  0.371 &
  0.663 &
  0.978 &
  0.458 &
  0.522 &
  \textbf{0.668} &
  0.905 &
  0.405 &
  \textbf{0.769} &
  0.626 &
  0.953 &
  0.420 &
  0.529 \\
 &
   &
   &
  Full PT &
  \textbf{0.587} &
  0.951 &
  \textbf{0.423} &
  0.377 &
  0.675 &
  0.964 &
  0.508 &
  0.504 &
  0.645 &
  0.910 &
  0.381 &
  0.724 &
  0.628 &
  0.948 &
  0.441 &
  0.510 \\ \cmidrule{2-20} 
 &
  \multirow{4}{*}{\begin{tabular}[c]{@{}l@{}}BM + \\ MSP\end{tabular}} &
  \multirow{2}{*}{BERT} &
  No PT &
  0.522 &
  0.946 &
  0.310 &
  0.308 &
  0.754 &
  0.989 &
  0.586 &
  0.694 &
  0.650 &
  0.939 &
  0.388 &
  0.690 &
  0.629 &
  0.960 &
  0.420 &
  0.532 \\
 &
   &
   &
  Full PT &
  0.488 &
  0.944 &
  0.246 &
  0.305 &
  0.756 &
  \textbf{0.997} &
  0.567 &
  0.713 &
  0.654 &
  0.950 &
  0.349 &
  0.739 &
  0.613 &
  \textbf{0.964} &
  0.374 &
  0.543 \\ \cmidrule{3-20} 
 &
   &
  \multirow{2}{*}{SciBERT} &
  No PT &
  0.527 &
  0.948 &
  0.314 &
  0.323 &
  \textbf{0.778} &
  0.992 &
  \textbf{0.628} &
  0.719 &
  0.655 &
  0.843 &
  \textbf{0.444} &
  0.728 &
  0.639 &
  0.945 &
  0.448 &
  0.554 \\
 &
   &
   &
  Full PT &
  0.538 &
  \textbf{0.952} &
  0.338 &
  0.317 &
  0.767 &
  0.992 &
  0.625 &
  0.669 &
  0.653 &
  0.899 &
  0.414 &
  0.691 &
  0.640 &
  0.957 &
  0.451 &
  0.526 \\ \cmidrule{2-20} 
 &
  \multirow{4}{*}{\begin{tabular}[c]{@{}l@{}}MSP + \\ Meas\\ Eval\end{tabular}} &
  \multirow{2}{*}{BERT} &
  No PT &
  0.556 &
  0.948 &
  0.374 &
  0.337 &
  0.746 &
  0.986 &
  0.561 &
  0.696 &
  0.523 &
  0.893 &
  0.187 &
  0.596 &
  0.614 &
  0.952 &
  0.398 &
  0.522 \\
 &
   &
   &
  Full PT &
  0.554 &
  0.939 &
  0.361 &
  0.362 &
  0.751 &
  \textbf{0.997} &
  0.576 &
  0.692 &
  0.500 &
  0.939 &
  0.185 &
  0.515 &
  0.610 &
  0.960 &
  0.396 &
  0.515 \\\cmidrule{3-20} 
 &
   &
  \multirow{2}{*}{SciBERT} &
  No PT &
  0.569 &
  0.948 &
  0.391 &
  0.360 &
  0.771 &
  \textbf{0.997} &
  0.609 &
  0.720 &
  0.562 &
  0.872 &
  0.241 &
  0.620 &
  0.639 &
  0.952 &
  0.441 &
  0.546 \\
 &
   &
   &
  Full PT &
  \textbf{0.594} &
  0.952 &
  \textbf{0.441} &
  \textbf{0.380} &
  0.759 &
  \textbf{0.997} &
  0.595 &
  0.691 &
  0.528 &
  0.872 &
  0.203 &
  0.635 &
  0.635 &
  0.954 &
  0.442 &
  0.549 \\ \midrule
\multirow{6}{*}{\begin{tabular}[c]{@{}c@{}}All \\ Sources\end{tabular}} &
  \multirow{6}{*}{\begin{tabular}[c]{@{}l@{}}MSP+ \\ Meas\\ Eval + \\ BM\end{tabular}} &
  \multirow{3}{*}{BERT} &
  No PT &
  0.568 &
  0.950 &
  0.391 &
  0.346 &
  0.762 &
  \textbf{0.997} &
  0.586 &
  0.703 &
  0.624 &
  0.939 &
  0.373 &
  0.637 &
  \textbf{0.646} &
  \textbf{0.965} &
  \textbf{0.454} &
  0.538 \\
 &
   &
   &
  Full PT &
  0.545 &
  0.938 &
  0.336 &
  0.351 &
  0.763 &
  \textbf{0.997} &
  0.576 &
  0.724 &
  0.661 &
  \textbf{0.961} &
  0.395 &
  0.695 &
  0.644 &
  \textbf{0.963} &
  0.430 &
  0.565 \\
 &
   &
   &
  Adapter PT &
  0.542 &
  0.946 &
  0.346 &
  0.342 &
  0.693 &
  0.989 &
  0.474 &
  0.644 &
  0.627 &
  0.927 &
  0.367 &
  0.675 &
  0.613 &
  0.958 &
  0.395 &
  0.527 \\ \cmidrule{3-20} 
 &
   &
  \multirow{3}{*}{SciBERT} &
  No PT &
  \textbf{0.586} &
  0.952 &
  0.389 &
  \textbf{0.412} &
  0.764 &
  0.992 &
  0.594 &
  0.716 &
  \textbf{0.676} &
  0.915 &
  0.417 &
  \textbf{0.754} &
  \textbf{0.667} &
  0.960 &
  \textbf{0.467} &
  \textbf{0.598} \\
 &
   &
   &
  Full PT &
  0.577 &
  0.943 &
  0.405 &
  0.365 &
  \textbf{0.779} &
  0.992 &
  \textbf{0.644} &
  0.687 &
  \textbf{0.670} &
  0.916 &
  0.412 &
  0.744 &
  \textbf{0.666} &
  0.956 &
  \textbf{0.485} &
  \textbf{0.577} \\
 &
   &
   &
  Adapter PT &
  0.550 &
  0.948 &
  0.363 &
  0.340 &
  0.751 &
  0.986 &
  0.559 &
  \textbf{0.733} &
  0.653 &
  0.883 &
  0.403 &
  0.733 &
  0.641 &
  0.950 &
  0.438 &
  \textbf{0.571} \\ \bottomrule
\end{tabular}%
}
\caption[Development F1 scores of Task 2]{Development F1 scores of Task 2: Context extraction. \textbf{Bold} scores indicate the highest score across an entire target domain. \textit{Italic} scores indicate the highest score within one source domain. O = Overall, U = Unit, ME = MeasuredEntity, MP = MeasuredProperty.}
\label{tab:dev_t2}
\end{table*}
\begin{table*}[!ht]
\centering
\resizebox{\textwidth}{!}{%
\begin{tabular}{llccccc|ccccc|ccccc|ccccc} \toprule
 &
   &
  \multicolumn{5}{c}{MeasEval} &
  \multicolumn{5}{c}{MSP} &
  \multicolumn{5}{c}{BM} &
  \multicolumn{5}{c}{Overall} \\\midrule
\begin{tabular}[c]{@{}l@{}}Source\\ domains\end{tabular} &
  Model configuration by Task &
  All &
  Q &
  U &
  ME &
  MP &
  All &
  Q &
  U &
  ME &
  MP &
  All &
  Q &
  U &
  ME &
  MP &
  All &
  Q &
  U &
  ME &
  MP \\ \midrule
\begin{tabular}[c]{@{}l@{}}MeasEval   \\ only\end{tabular} &
  \begin{tabular}[c]{@{}l@{}}T1: SciBERT Full PT; \\ T2: SciBERT Full PT\end{tabular} &
  0.555 &
  0.840 &
  0.847 &
  0.317 &
  0.312 &
  0.663 &
  0.872 &
  0.909 &
  0.472 &
  0.427 &
  0.394 &
  0.375 &
  0.534 &
  0.285 &
  0.416 &
  0.539 &
  0.687 &
  0.785 &
  0.354 &
  0.380 \\ \midrule
\begin{tabular}[c]{@{}l@{}}MSP\\ only\end{tabular} &
  \begin{tabular}[c]{@{}l@{}}T1: SciBERT No PT; \\ T2: SciBERT No PT\end{tabular} &
  0.502 &
  0.777 &
  0.846 &
  0.262 &
  0.218 &
  0.803 &
  0.929 &
  0.958 &
  0.668 &
  0.649 &
  0.340 &
  0.377 &
  0.446 &
  0.203 &
  0.374 &
  0.529 &
  0.681 &
  0.769 &
  0.341 &
  0.378 \\ \midrule
\begin{tabular}[c]{@{}l@{}}BM \\ only\end{tabular} &
  \begin{tabular}[c]{@{}l@{}}T1: SciBERT Full PT; \\ T2: SciBERT Full PT\end{tabular} &
  0.342 &
  0.543 &
  0.582 &
  0.115 &
  0.209 &
  0.188 &
  0.298 &
  0.215 &
  0.140 &
  0.076 &
  0.628 &
  0.775 &
  0.718 &
  0.416 &
  0.677 &
  0.357 &
  0.512 &
  0.476 &
  0.185 &
  0.302 \\ \midrule
Multi &
  \begin{tabular}[c]{@{}l@{}}T1: BM SciBERT Full PT \& \\ MSP+MeasEval Full PT; \\ T2: All Sources  SciBERT No PT\end{tabular} &
  0.647 &
  0.782 &
  0.893 &
  0.445 &
  0.456 &
  0.776 &
  0.930 &
  0.957 &
  0.532 &
  0.688 &
  0.450 &
  0.505 &
  0.519 &
  0.354 &
  0.440 &
  0.641 &
  0.767 &
  0.848 &
  0.450 &
  0.505 \\ \bottomrule
\end{tabular}%
}
\caption{Development E2E (strict) F1}
\label{tab:dev_e2e}
\end{table*}

\clearpage

\section*{J \quad Re-Annotation guidelines for the Material Synthesis Procedural Text Corpus}\label{appendix-guidelines}
Scientific knowledge is published and achieved in the form of unstructured texts. Numeric components in the form of counts, measurements and units (e.g. 500mg) and their contexts (e.g. Ibuprofen, dosage) are often crucial information for researchers across all domains. The goal of this annotation task is to prepare data for an end-to-end pipeline, which is able to extract quantities, units, measured objects and properties from texts, as well as the semantic relationships between each other. Section A) of this document describes how the entity and relation labels can be defined in a general setting. Section B) will provide guidance for the annotation of a specific dataset, the \href{https://github.com/olivettigroup/annotated-materials-syntheses}{Materials Science Procedural Text Corpus by Mysore et al.}.
\section*{A) General task guidelines}
The entity and relation labels are described in the following table. They are a subset of the \href{https://github.com/harperco/MeasEval/tree/main/annotationGuidelines#basic-annotation-set}{SemEval 2021 Task 8, MeasEval Basic Annotation Set}.

\begin{itemize}
    \item \textbf{Number (N)} 
    \begin{itemize}
        \item \textbf{Definition}A numeric value or a count signifying an amount or measurement and contiguous specifiers (e.g. >, \~{}). This is the root entity in each sample, i.e. other entities must always be able to directly refer to a number. Numeric values which do not signify a quantifiable amount (e.g. page numbers, citations, mathematical formulas) are not annotated.
        \item \textbf{Example} The patient weighted \textbf{\~{}100} pounds and was prescribed an Ibuprofen dosage of \textbf{500} mg. 
    \end{itemize}
    \item \textbf{Unit (U) } 
    \begin{itemize}
        \item \textbf{Definition} The unit linked to the Number. To be annotated if available. 
        \item \textbf{Example} The sick patient weighted 100 \textbf{pounds} and was prescribed an Ibuprofen dosage of 500 \textbf{mg}. 
    \end{itemize}
    \item \textbf{measuredEntity (mE)} 
    \begin{itemize}
        \item \textbf{Definition} "A required (if possible) span that has a given [Number + Unit] either as its direct value or indirectly via a MeasuredProperty. Every Quantity should ideally be associated with a MeasuredEntity. If no relevant information appears in the text, the Number can be standalone, but can have no other relationships. A MeasuredEntity can be related to either a MeasuredProperty by a HasProperty relationship, or to a Quantity by a HasQuantity relationship." (cited from SemEval 2021 Task 8 \href{https://github.com/harperco/MeasEval/tree/main/annotationGuidelines}{annotation guidelines}, "Quantity" reference replaced with "Number"). This label describes the concept that is being quantified by the number (and the unit). In most cases the measuredEntity consists of one or more noun phrases (and their specifiers if they are in a contiguous span).
        \item \textbf{Example}  The \textbf{sick patient} weighted \~{}100 pounds and was prescribed an \textbf{Ibuprofen} dosage of 500 mg.  
    \end{itemize}
    \item \textbf{measuredProperty (mP)} 
    \begin{itemize}
        \item \textbf{Definition} "An optional span associated with both a MeasuredEntity and a [Number]. Not every [Number] will be associated with a MeasuredProperty. A MeasuredProperty must be related from a MeasuredEntity by a HasProperty relationship, and must be related to a Quantity through the HasQuantity relationship." (cited from SemEval 2021 Task 8 \href{https://github.com/harperco/MeasEval/tree/main/annotationGuidelines}{annotation guidelines}, "Quantity" reference replaced with "Number"). The measuredProperty can be interpreted as the "quantity-denoting target-word" of the number (definition from \href{https://framenet2.icsi.berkeley.edu/fnReports/data/frameIndex.xml?frame=Quantity}{FrameNet}). As such is it often a quantifiable specifier or attribute of the measuredEntity (e.g. volume, concentration, temperature etc.), but can also encompass longer target phrases.
        \item \textbf{Example} The patient \textbf{weighted} \~{}100 pounds and was prescribed an Ibuprofen \textbf{dosage} of 500 mg.
    \end{itemize}
    
\end{itemize}

\noindent \textbf{Graph representation}: The entity labels their relations can be depicted in a graph. This can be especially helpful when identifying the measured entity and measured property or verifiying one's annotations.

Case 1: (N, U) <- hasQuantity <- (mP) <- hasProperty <- (mE)

Case 2: (N, U) <- hasQuantity <- (mE)

\noindent Each data sample must contain at least a Number. The other labels are only to be annotated if they are contained in the text. Below are a few hints and rules for the general annotation task. Quantity and the Number label will be used synonymously.

\subsection*{A.1) Multi-class classification}
Multi-entity classifications are possible, i.e. a measuredEntity for one Number can be a measuredProperty for another. This can be the case, because classification is always performed from the perspective of the root Number. For the same reason there can be measuredEntities or measuredProperties containing numbers (that are not the root number of the annotation sample).

\noindent \textbf{Examples}

\begin{displayquote}
"The lowest input of odd nitrogen corresponds to 3.5-6.1 (x10-4) wt.\% N accumulated over 3 byr and mixed into 1.5-2.6 m, of soil."
\end{displayquote}

\begin{table}[!h]
\resizebox{\textwidth}{!}{%
\begin{tabular}{|c|c|c|c|}
\hline
N & U & mE & mP\\
\hline
3.5-6.1 (x10-4) & wt.\% & N & lowest input of odd nitrogen \\
\hline
\end{tabular}}
\end{table}

\begin{table}[!h]
\resizebox{\textwidth}{!}{%
\begin{tabular}{|c|c|c|c|}
\hline
N & U & mE & mP\\
\hline
3 & byr & lowest input of odd nitrogen & accumulated \\
\hline
\end{tabular}}
\end{table}

\begin{table}[!h]
\begin{tabular}{|c|c|c|c|}
\hline
N & U & mE & mP\\
\hline
1.5-2.6 & m & soil &  \\
\hline
\end{tabular}
\end{table}

\subsection*{A.2) Span extent}
We annotate measuredEntities and measuredProperties as completely as possible, i.e. using the longest coherent and informative text span. However, we do not annotate copula (e.g. were, have been \href{https://en.wikipedia.org/wiki/Copula_(linguistics)}{etc.}) prepositions or articles at the beginning or end of a span.

\noindent \textbf{Examples}

\begin{displayquote}
"The earth surface temperatures have risen by 0.5 °C compared to baseline levels."
\end{displayquote}

\begin{table}[!h]
\begin{center}
\begin{tabular}{|c|c|c|c|}
\hline
N & U & mE & mP\\
\hline
0.5 & °C & earth surface temperatures & risen \\
\hline
\end{tabular}
\end{center}
\end{table}

\subsection*{A.3) Duplicate measuredEntities mentions}
Some sentences will have multiple mentions of the same measuredEntity. We annotate the span that is closest to its root Number.

\noindent \textbf{Example}

\begin{displayquote}
"The O2/N ratio was measured with the aforementioned machinery (O2/N = 2.8)."
\end{displayquote}

\begin{table}[!h]
\begin{center}
\begin{tabular}{|c|c|c|c|}
\hline
N & U & mE & mP\\
\hline
2.8 &  & O2/N & ratio \\
\hline
\end{tabular}
\end{center}
\end{table}

\subsection*{A.4) Part-whole relationships}
Fractions and percentages often describe part-whole relationships, where the fraction or percentage describe a partial characteristic of a bigger whole. For annotation, we mark the whole as the measuredEntity and the part as the measuredProperty.

\noindent \textbf{Examples}

\begin{displayquote}
"The hamburger consisted of 30\% patty and 10\% cheese."
\end{displayquote}

\begin{table}[!h]
\begin{center}
\begin{tabular}{|c|c|c|c|}
\hline
N & U & mE & mP\\
\hline
30 & \% & hamburger & patty \\
\hline
\end{tabular}
\end{center}
\end{table}

\begin{table}[!h]
\begin{center}
\begin{tabular}{|c|c|c|c|}
\hline
N & U & mE & mP\\
\hline
10 & \% & hamburger & cheese \\
\hline
\end{tabular}
\end{center}
\end{table}

\begin{displayquote}
"Steam activation was carried out by heating an amount of sample in a flow of 10\% water vapor."
\end{displayquote}

\begin{table}[!h]
\begin{center}
\begin{tabular}{|c|c|c|c|}
\hline
N & U & mE & mP\\
\hline
10 & \% & flow & water vapor \\
\hline
\end{tabular}
\end{center}
\end{table}

\noindent Part-whole relationships can also be described without the use of fractions or percentages:

\begin{displayquote}
"The patty of the hamburger was 200g."
\end{displayquote}

\begin{table}[!h]
\begin{center}
\begin{tabular}{|c|c|c|c|}
\hline
N & U & mE & mP\\
\hline
200 & g & hamburger & patty \\
\hline
\end{tabular}
\end{center}
\end{table}

\noindent Graph representation: hamburger -> hasProperty -> patty -> hasQuantity -> 200g

\subsection*{A.5) Non-noun measuredProperties}
MeasuredProperties can also be verbs or adjectives. To test whether a verb can be a measuredProperty, one can reformulate the sentence using the nominalized verb form valditate with the graph representation to check if all relations can be applied correctly.

\noindent \textbf{Examples}

\begin{displayquote}
"The earth surface temperatures have risen by 0.5 °C compared to baseline levels."
\end{displayquote}

\noindent Reformulated: There has been a rise of earth surface temperature by 0.5 °C compared to baseline levels.

\begin{displayquote}
"The patient weighed 100 pounds."
\end{displayquote}

\noindent Reformulated: The weight of the patient is 100 pounds.

\subsection*{A.6) Hints for the Unit entity}
Ratios (e.g. weight ratio) and pH values are not considered units. Instead, they are labeled as measuredProperties.

\section*{B) Specific guidelines for Mysore et al.'s Materials Science Procedural (MSP) Text Corpus}
Originally, the MSP Corpus contains annotations regarding the materials, operations and conditions of experiments in materials science.\\
To expand the existing MeasEval Dataset, we need to adapt these annotations to the above-introduced entities and labels.\\
For this, the dataset was automatically processed beforehand using mapping rules for each pre-existing label (e.g. all materials where labeled as measuredEntities). However, these automatically created labels are often incorrect and must be adjusted which is the main annotation task here.

\textbf{Characteristics of the data:}\\
Each data sample is pre-labeled with at least a number and in most cases suggestions for the Unit, measuredProperty and measuredEntity are given.\\
One data sample is created for each Number and its related measuredEntities and measuredProperties. Hence, sentences with multiple quantities and related contexts will yield as many data samples as there are Numbers in the text. Due to the specificity of this corpus, some additional rules apply. They are listed below.

\subsection*{B.1) Number specifiers}
Symbols and textual specifiers of Numbers are often not included in the label suggestion. Therefore, we must expand the Number-span to also contain these specifiers.
Example:
In the first example the suggested number would be "The patient weighted \~{} \textbf{[100]} pounds...", we would then extend the span to "The patient weighted \textbf{[\~{}100]} pounds...".

\subsection*{B.2) Removing irrelevant entities and adjusting spans}
Sometimes the there will be suggested entities, that are not related to the root Number. These false suggestions must be removed.\\
Further, to adhere to the rule for maximum span annotation we adjust spans for measuredEntities and measuredProperties which can be extended.

\noindent \textbf{Examples}

\begin{displayquote}
"The gel was ground to powders and then calcined at 400 °C in a muffle furnace under air atmosphere."
\end{displayquote}

\noindent Suggested:

\begin{table}[!h]
\begin{center}
\begin{tabular}{|c|c|c|c|}
\hline
N & U & mE & mP\\
\hline
400 & °C & muffle furnace, air & calcinated \\
\hline
\end{tabular}
\end{center}
\end{table}

\noindent Corrected:

\begin{table}[!h]
\begin{center}
\begin{tabular}{|c|c|c|c|}
\hline
N & U & mE & mP\\
\hline
400 & °C & powders & calcinated \\
\hline
\end{tabular}
\end{center}
\end{table}

\begin{displayquote}
"The as-synthesized zeolites were calcined at 580 degC for 4 h under a flow of air."
\end{displayquote}

\noindent Suggested measuredEntity = zeolites \\
\noindent Corrected measuredEntity = as-synthesized zeolites

\subsection*{B.3) Experiment procedures}
A large fraction of this corpus' Numbers describe experimental conditions, e.g. how long a solution was stirred. As a result, the quantities can often not be linked to an explicitly measured object, but only to the object that is being experimented on. Therefore, we mark these objects as measuredEntities and the experimental circumstances as measuredProperties. If the measuredEntity is explicitly given, we mark that as measuredEntity instead of the object that is impacted by the experiment.

\noindent \textbf{Examples}

\begin{displayquote}
"The obtained sample was washed with absolute ethanol, and then dried at 60 °C for 10h."
\end{displayquote}

\begin{table}[!h]
\begin{center}
\begin{tabular}{|c|c|c|c|}
\hline
N & U & mE & mP\\
\hline
60 & °C & obtained sample & dried \\
\hline
\end{tabular}
\end{center}
\end{table}

\begin{table}[!h]
\begin{center}
\begin{tabular}{|c|c|c|c|}
\hline
N & U & mE & mP\\
\hline
10 & h & obtained sample & dried \\
\hline
\end{tabular}
\end{center}
\end{table}

\begin{displayquote}
"The solution was modified by dissolving it in 10 wt\% ethanol."
\end{displayquote}

\begin{table}[!h]
\begin{center}
\begin{tabular}{|c|c|c|c|}
\hline
N & U & mE & mP\\
\hline
10 & wt\% & ethanol &  \\
\hline
\end{tabular}
\end{center}
\end{table}

\subsubsection*{B.3.1) Experiment operations}
We only mark procedural operations (e.g. added or dissolved) as the measuredProperty of a Number and a measuredEntity, if the measuredEntity is the main participant of the operation.

\noindent \textbf{Examples}

\begin{displayquote}
"The solution was modified by dissolving it in 10 wt\% ethanol."
\end{displayquote}

\noindent In the example above we do not mark \textit{dissolving} as the mP, because ethanol is not the component that is being dissolved.

\begin{displayquote}
"500 g of the sample was dissolved in 10 ml NaCl solution."
\end{displayquote}

\begin{table}[!h]
\begin{center}
\begin{tabular}{|c|c|c|c|}
\hline
N & U & mE & mP\\
\hline
500 & g & sample & dissolved \\
\hline
\end{tabular}
\end{center}
\end{table}

\begin{table}[!h]
\begin{center}
\begin{tabular}{|c|c|c|c|}
\hline
N & U & mE & mP\\
\hline
10 & ml & NaCL solution &  \\
\hline
\end{tabular}
\end{center}
\end{table}

\noindent Here the sample is the entity that is being dissolved, thus we can mark 'dissolved' as its measuredProperty. Be careful that the operation marked as the measuredProperty has a proper relation to the Number span.

\begin{displayquote}
"The composite was ground, pressed and sintered at 300 °C."
\end{displayquote}

\begin{table}[!h]
\begin{center}
\begin{tabular}{|c|c|c|c|}
\hline
N & U & mE & mP\\
\hline
300 & °C & composite & sintered \\
\hline
\end{tabular}
\end{center}
\end{table}

\noindent In this example, sintered is the operation which directly related to the temperature measure, whereas the other operations \textbf{do no}t have a measuredProperty (e.g. what the pressure of the pressing was or how granular the grounding was).

\begin{displayquote}
"Copper (99,99\%) was purchased from Sigma-Aldrich."
\end{displayquote}

\noindent In this case 'purchased' is not the target-word of the 99,99\%, as it represents a purity measure and not an amount that was purchased.\\
Thus, there is no measuredProperty in this sentence.

\subsection*{B.3.2) MeasuredProperty operations span}
Operations are often specified by additional descriptors, that are contiguous to the operation or in a separate span. In most cases we only annotate the operation as the measuredProperty, because the descriptors are semantically dependent (the so-called \href{https://universaldependencies.org/u/dep/obl.html}{'oblique nominal'}) on the operation phrase, which is difficult to express within our annotation scheme.

\noindent \textbf{Examples}

\begin{displayquote}
"The chemical was heated at 300 °C under constant airflow."
\end{displayquote}

\begin{table}[!h]
\begin{center}
\begin{tabular}{|c|c|c|c|}
\hline
N & U & mE & mP\\
\hline
300 & °C & chemical & heated \\
\hline
\end{tabular}
\end{center}
\end{table}

\noindent Here, "under constant airflow" is dependent on "heated". We would need additional labels to capture these kind of multi-level relations, which exceeds the scope of this annotation scheme. In the case, that a contiguous span with multiple properties could be annotated, we proceed in the same manner and only annotate the highest level to stay consistent.

\begin{displayquote}
"The chemical was dried in air at 300 °C."
\end{displayquote}

\begin{table}[!h]
\begin{center}
\begin{tabular}{|c|c|c|c|}
\hline
N & U & mE & mP\\
\hline
300 & °C & chemical & dried \\
\hline
\end{tabular}
\end{center}
\end{table}

\noindent \textbf{Exception: Preceding adverbial and adjectival modifiers}\\
We can add such descriptors which occur prior to the operation to the operation measuredProperty span, as they almost exclusively occur together contiguously, thus ensuring consistent annotations.

\begin{displayquote}
"The chemical was under magnetic stirring for 2 h."
\end{displayquote}

\begin{table}[!h]
\begin{center}
\begin{tabular}{|c|c|c|c|}
\hline
N & U & mE & mP\\
\hline
2 & h & chemical & magnetic stirring \\
\hline
\end{tabular}
\end{center}
\end{table}

\subsubsection*{B.3.3) Properties of operations}
When attributes or specifiers of an operation are given we try to mark the main experiment participant as the measuredProperty as opposed to the operation itself.

\noindent \textbf{Example}

\begin{displayquote}
"The chemical was calcinated at 300 °C with a heating rate of 10 °C per minute."
\end{displayquote}

\begin{table}[!h]
\begin{center}
\begin{tabular}{|c|c|c|c|}
\hline
N & U & mE & mP\\
\hline
10 & °C per minute & chemical & heating rate \\
\hline
\end{tabular}
\end{center}
\end{table}

\subsection*{B.4) Ambivalent relations}
Some experimental descriptions do not explicitly name the measuredEntity, but e.g. only the result of the experimental operation. If this is the case and an experimental operation is also in the sentence, we can mark the experimental operation as the measuredProperty.

\noindent \textbf{Examples}

\begin{displayquote}
"The enhanced form was obtained by calcination at 220 °C under a flow of air."
\end{displayquote}

\begin{table}[!h]
\begin{center}
\begin{tabular}{|c|c|c|c|}
\hline
N & U & mE & mP\\
\hline
220 & °C & calcination & under a flow of air \\
\hline
\end{tabular}
\end{center}
\end{table}

\noindent This sentence does not mention the object that is being calcinated. Therefore, we annotate the operation as the measuredEntity.\\
Note that we can also annotate "under a flow of air" as a measuredProperty here, because it can be directly linked to "calcination" and "220 °C" without being dependent on another measuredProperty.

\begin{displayquote}
"NH4OH solution was slowly added until the pH was 10."
\end{displayquote}

\begin{table}[!h]
\begin{center}
\begin{tabular}{|c|c|c|c|}
\hline
N & U & mE & mP\\
\hline
10 &  & pH &  \\
\hline
\end{tabular}
\end{center}
\end{table}

\noindent This sentence does not mention, whose pH becomes 10. Because pH is not a unit, we can mark it as the measuredEntity.

\subsubsection*{B.4.1) Coreferences}
If the measuredEntity is mentioned as a coreference, but not explicitly, we annotate the coreference as measuredEntity.

\noindent \textbf{Example}

\begin{displayquote}
"Finally, it was filtered, washed with water and ethanol, and vacuum-dried at 70 °C."
\end{displayquote}

\begin{table}[!h]
\begin{center}
\begin{tabular}{|c|c|c|c|}
\hline
N & U & mE & mP\\
\hline
70 & °C & it & vacuum-dried \\
\hline
\end{tabular}
\end{center}
\end{table}

\subsubsection*{B.4.2) Transformation of the measuredEntity}
Synthesis procedures often describe transformations of the measuredEntities before a measurable operation occurs. It is often not possible to pin-point one particular noun phrase that represents the entity to which the operation is being applied. Instead, we annotate all prior steps that are relevant for the operation as the measuredEntity.

\noindent \textbf{Examples}

\begin{displayquote}
"To prepare C3N4-Pd composites, the as-prepared g-C3N4 was added into 100 mL ethanol and was sonicated for 2 h to obtain thin g-C3N4 nanosheets."
\end{displayquote}

\begin{table}[!h]
\begin{center}
\resizebox{\textwidth}{!}{
\begin{tabular}{|c|c|c|c|}
\hline
N & U & mE & mP\\
\hline
2 & h & as-prepared g-C3N4 was added into 100 mL ethanol & sonicated \\
\hline
\end{tabular}
}
\end{center}
\end{table}

\begin{table}[!h]
\begin{center}
\begin{tabular}{|c|c|c|c|}
\hline
N & U & mE & mP\\
\hline
100 & mL & ethanol &  \\
\hline
\end{tabular}
\end{center}
\end{table}

\subsection*{B.5) Dealing with nested information in brackets}
Sometimes additional information about a measuredEntity is given in brackets. We only annotate measuredProperties that are directly related to both the Number and the measuredEntity.

\noindent \textbf{Examples}

\begin{displayquote}
"20g of gold (99.99\% purity) were ground."
\end{displayquote}

\begin{table}[!h]
\begin{center}
\begin{tabular}{|c|c|c|c|}
\hline
N & U & mE & mP\\
\hline
20 & g & gold & ground \\
\hline
\end{tabular}
\end{center}
\end{table}

\begin{table}[!h]
\begin{center}
\begin{tabular}{|c|c|c|c|}
\hline
N & U & mE & mP\\
\hline
99.99 & \% & gold & purity \\
\hline
\end{tabular}
\end{center}
\end{table}

\begin{displayquote}
"In a typical process, NiCl2*6H2O (0.173 g) was dissolved in a solution."
\end{displayquote}

\begin{table}[!h]
\begin{center}
\begin{tabular}{|c|c|c|c|}
\hline
N & U & mE & mP\\
\hline
0.173 & g & NiCl2*6H2O & dissolved \\
\hline
\end{tabular}
\end{center}
\end{table}

\subsection*{B.6) MeasuredEntity for Ratios}
Ratios explain "how many times one number contains another" (\href{https://en.wikipedia.org/wiki/Ratio}{Wiki}). This should also be expressed in the measuredEntity of a ratio. If the two concepts described by the ratio are explicitly mentioned, annotate them (either in a contiguous span if possible, and separately if not).

\noindent \textbf{Examples}

\begin{displayquote}
"At a weight ratio of 1:1, the MWCNT@MPC composite was mixed with sublimed sulfur."
\end{displayquote}

\begin{table}[!h]
\begin{center}
\resizebox{\textwidth}{!}{%
\begin{tabular}{|c|c|c|c|}
\hline
N & U & mE & mP\\
\hline
1:1 &  & MWCNT@MPC composite was mixed with sublimed sulfur & weight ratio \\
\hline
\end{tabular}
}
\end{center}
\end{table}

\subsection*{B.7) Abbreviations}
\begin{table}[!h]
\begin{center}
\begin{tabular}{|c|c|c|c|}
\hline
N & U & mE & mP\\
\hline
100 & ml & Hydrochloric acid (HCl & added \\
\hline
\end{tabular}
\end{center}
\end{table}

We try to include abbreviations into the entity span, if possible.

\noindent \textbf{Example}

\begin{displayquote}
"Hydrochloric acid (HCl, 100 ml) was added to the mixture."
\end{displayquote}

\end{document}